\documentclass[lettersize,journal]{IEEEtran}
\usepackage{amsmath,amsfonts}
\usepackage{algorithmic}
\usepackage{algorithm}
\usepackage{array}
\usepackage{subfig}
\usepackage{float}
\usepackage{textcomp}
\usepackage{stfloats}
\usepackage{url}
\usepackage{verbatim}
\usepackage{graphicx}
\usepackage{cite}
\usepackage{verbatim}
\usepackage{graphicx}
\usepackage{threeparttable}
\usepackage{makecell}
\usepackage{bm}
\usepackage{booktabs} 
\usepackage{multirow}
\usepackage{adjustbox}
\usepackage{hyperref}
\usepackage{color}
\usepackage{url}            
\begin{document}

\title{\textcolor{red}{Bi-level Multi-objective Evolutionary Learning: A Case Study on Multi-task Graph Neural Topology Search}}

\author{Chao Wang, Licheng Jiao,~\IEEEmembership{Fellow,~IEEE, }Jiaxuan Zhao,~\IEEEmembership{Student Member,~IEEE, }Lingling Li,~\IEEEmembership{Senior Member,~IEEE}, Xu Liu,~\IEEEmembership{Member,~IEEE}, Fang Liu,~\IEEEmembership{Senior Member,~IEEE}, Shuyuan Yang,~\IEEEmembership{Senior Member,~IEEE}
\thanks{This work was supported in part by the Key Scientific Technological Innovation Research Project by Ministry of Education, the National Natural Science Foundation of China Innovation Research Group Fund (61621005), the State Key Program and the Foundation for Innovative Research Groups of the National Natural Science Foundation of China (61836009), the Major Research Plan of the National Natural Science Foundation of China (91438201, 91438103, and 91838303), the National Natural Science Foundation of China (U1701267, 62076192, 62006177, 61902298, 61573267, 61906150, and 62276199), the 111 Project, the Program for Cheung Kong Scholars and Innovative Research Team in University (IRT 15R53), the ST Innovation Project from the Chinese Ministry of Education, the Key Research and Development Program in Shaanxi Province of China (2019ZDLGY03-06), the National Science Basic Research Plan in Shaanxi Province of China (2019JQ-659, 2022JQ-607), China Postdoctoral fund (2022T150506) the Scientific Research Project of Education Department In Shaanxi Province of China (No.20JY023), the fundamental research funds for the central universities (XJS201901, XJS201903, JBF201905, JB211908), and the CAAI-Huawei MindSpore Open Fund. (Corresponding author: Licheng Jiao.)}
\thanks{The authors are with the Key Laboratory of Intelligent Perception and Image Understanding of Ministry of Education, International Research Center for Intelligent Perception and Computation, Xidian University, Xi’an 710071, China (e-mail: lchjiao@mail.xidian.edu.cn).}}

\markboth{~Vol.~XXX, No.~XXX, January~2023}%
{Shell \MakeLowercase{\textit{et al.}}: A Sample Article Using IEEEtran.cls for IEEE Journals}


\maketitle

\begin{abstract}
The construction of machine learning models involves many bi-level multi-objective optimization problems (BL-MOPs), where upper level (UL) candidate solutions must be evaluated via training weights of a model in the lower level (LL). Due to the Pareto optimality of sub-problems and the complex dependency across UL solutions and LL weights, an UL solution is feasible if and only if the LL weight is Pareto optimal. It is computationally expensive to determine which LL Pareto weight in the LL Pareto weight set is the most appropriate for each UL solution. This paper proposes a bi-level multi-objective learning framework (BLMOL), coupling the above decision-making process with the optimization process of the UL-MOP by introducing LL preference $\bm{r}$. Specifically, the UL variable and $\bm{r}$ are simultaneously searched to minimize multiple UL objectives by evolutionary multi-objective algorithms. The LL weight with respect to $\bm{r}$ is trained to minimize multiple LL objectives via gradient-based preference multi-objective algorithms. In addition, the preference surrogate model is constructed to replace the expensive evaluation process of the UL-MOP. We consider a novel case study on multi-task graph neural topology search. It aims to find a set of Pareto topologies and their Pareto weights, representing different trade-offs across tasks at UL and LL, respectively. The found graph neural network is employed to solve multiple tasks simultaneously, including graph classification, node classification, and link prediction. Experimental results demonstrate that BLMOL can outperform some state-of-the-art algorithms and generate well-representative UL solutions and LL weights.
\end{abstract}

\begin{IEEEkeywords}
Bi-level multi-objective learning, multi-task learning, neural topology search.
\end{IEEEkeywords}

\section{Introduction}
\IEEEPARstart{T}{here} are multiple non-convex, non-differentiable, or even undefinable optimization objectives in a machine learning system, which are conveniently modeled as bi-level multi-objective optimization problems (BL-MOPs), such as neural architecture search (NAS)\cite{lu2020nsganetv2,lu2021neural}, hyper-parameter optimization (HPO)\cite{9744035,Belakaria_Deshwal_Doppa_2020}, and meta-learning \cite{ye2021multi}. Each level of these problems is a sub-problem with multiple complex objectives and variables\cite{9766417}. In a machine learning system, the upper level MOP (UL-MOP) deals with a number of explorable components (variables), and the lower level MOP (LL-MOP) involves training weights of a model, which are interdependent and together determine the form of UL and LL objectives. For example, the BL-MOP optimizes the UL architecture and LL weight to minimize multiple objectives at both levels for NAS problems. UL objectives are often non-differentiable or even black-box, such as the number of parameters and inference delay. And LL objectives are high-dimensional, differentiable, and non-convex, such as error rate and model sparsity.

\begin{figure*}[htbp]
\centering
\includegraphics[width=0.85\textwidth]{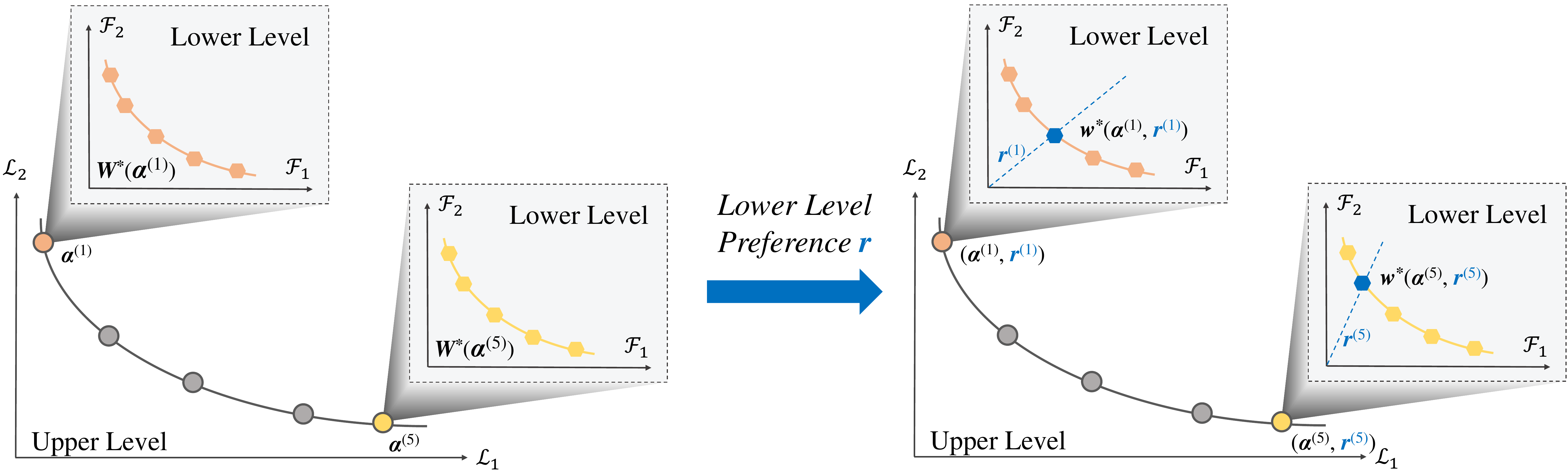}
\caption{Bi-level multi-objective optimization problem in machine learning (Left). The decision-making process of the LL-MOP is coupled to the UL-MOP by introducing LL preference $\bm{r}$ (Right).} \label{fig_1}
\end{figure*}

As shown in Fig. \ref{fig_1} (Left), because of multiple conflicting objectives for each sub-problem, it is likely that there will be multiple Pareto optimal solutions at each level. In addition, given an UL solution $\bm{\alpha}^{(1)}$, the LL-MOP aims to find a LL Pareto weight set $\bm{W^*}(\bm{\alpha}^{(1)})$. Therefore, each UL solution is associated with a LL Pareto weight set. We expect to select a LL Pareto weight from the LL Pareto weight set that makes the overall solutions optimal at the UL, called the \textit{\textbf{decision-making process of the LL-MOP}}\cite{9766417}. Due to requiring multiple LL Pareto weights to be found for an UL solution, the decision-making process is computationally expensive, which poses a significant challenge in deep learning.

To avoid the above issues, a common intuition is directly approximating the LL-MOP as a single-objective problem\cite{ye2021multi,franceschi2018bilevel}, which minimizes a weighted linear combination of LL objectives (losses). However, the performance of this workaround is severely affected by the manually tuned loss weights when LL losses may conflict. In addition, there are many excellent works using population-based algorithms to solve the low-dimensional BL-MOPs \cite{9766417, deb2009solving, ruuska2012constructing, sinha2011bilevel}. However, since maintaining multiple high-dimensional LL weights \cite{ye2021multi} is computationally expensive, these methods are not directly applicable to BL-MOPs in machine learning, especially in deep learning.

In this paper, for the first time, we propose a bi-level multi-objective learning framework (BLMOL) to solve BL-MOPs in machine learning as shown in Fig. \ref{fig_1} (Right). First, BLMOL couples the decision-making process of the LL-MOP to the UL-MOP by introducing LL preference $\bm{r}$. In particular, the UL variable $\bm{\alpha}$ and the LL preference $\bm{r}$ are simultaneously searched to minimize multiple UL objectives by multi-objective evolutionary algorithms (MOEAs). For an UL solution $\bm{\alpha}^{(1)}$ and a LL preference $\bm{r}^{(1)}$, the LL-MOP is transformed into a preference LL-MOP to find a LL Pareto optimal weight $\bm{w^*}(\bm{\alpha}^{(1)}, \bm{r}^{(1)})$ with preference $\bm{r}$. Gradient-based preference multi-objective algorithms (GPMOAs) can be employed to solve the preference LL-MOP. In practical machine learning systems, it is expensive to evaluate each UL solution at the UL in real time. This is because training a LL weight with preference $\bm{r}$ for a candidate UL solution by GPMOAs necessitates multiple costly iterations of gradient descent over multiple epochs. Therefore, the computationally cheap preference surrogate model is constructed using a certain number of expensive function evaluations to replace multiple UL objective functions during the evaluation process, whose input is the UL variable and the LL preference $\bm{r}$ and output is the predicted value of the UL-MOP. Finally, we can choose the specific multi-objective optimizer according to the functional properties of the BL-MOP.

While being applicable to various BL-MOPs in machine learning, our research takes multi-task graph neural topology search (MTGNTS) as a typical example. Graph neural network (GNN) \cite{xu2018how} captures the graph information by message passing with the neighbors on the graph data to produce highly task-specific embeddings in an end-to-end manner. Due to the low transferability of embeddings and limited representational capabilities of handcrafted GNNs, existing work cannot efficiently handle multiple downstream graph tasks simultaneously, such as graph classification (GC), node classification (NC), and link prediction (LP) \cite{cai2018comprehensive}. The topology of GNNs has a substantial effect representational ability of the model\cite{10.1145/3485447.3512185}. Manually designing an ideal multi-task GNN (MTGNN) topology that can learn the graph embedding suitable for multiple tasks is time-consuming and labor-intensive. Since tasks (objectives) may conflict, users are forced to make trade-offs among them. Keeping these in mind, we ingeniously consider automatically searching for general-purpose MTGNN topologies that match various user-desired task preferences, which is naturally modeled as a BL-MOP.

\begin{figure*}[htbp]
\centering
\includegraphics[width=0.85\textwidth]{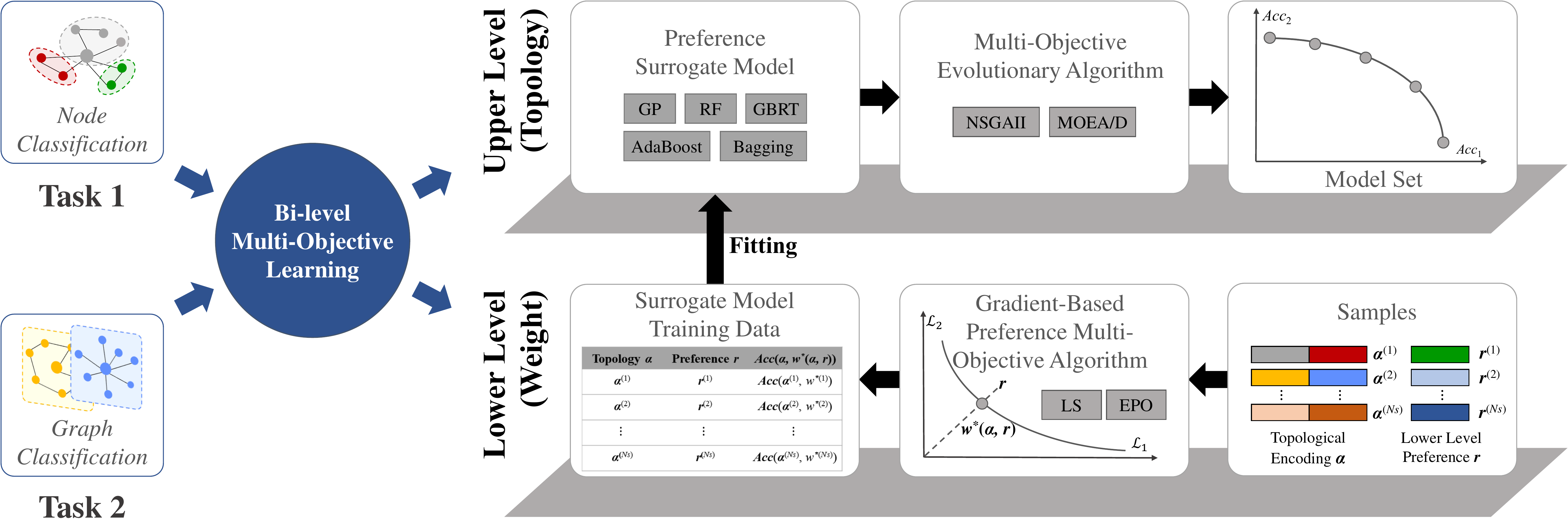}
\caption{Outline of BLMOL (Taking MTGNTS for the node classification and the graph classification as an example).} \label{fig_2}
\end{figure*}

In MTGNTS, the BL-MOP is intended to find multiple UL Pareto topologies and their LL Pareto weights. The UL and LL objectives are multiple task accuracy rates on the validation data and multiple task losses on the training data, respectively. Therefore, as shown in Fig. \ref{fig_2}, we adopt our proposed BLMOL to solve the MTGNTS problem. First, some topologies $\bm{\alpha}$ and LL preference vectors $\bm{r}$ are sampled from the search space. For each sample pair $(\bm{\alpha}^{(j)},\bm{r}^{(j)})$, the LL weight $\bm{w}^{(j)}$ is trained by GPMOAs over multiple epochs, and then multiple accuracy rates $\bm{Acc}(\bm{\alpha}^{(j)},\bm{w}^{\bm{*}(j)})$ are calculated. Next, preference surrogate models are developed to fit task accuracy rates $\bm{Acc}(\bm{\alpha},\bm{w^*}(\bm{\alpha},\bm{r}))$. Finally, MOEAs are employed to search the topologies and LL preference vectors to maximize task accuracy rates. To improve search efficiency, the aforementioned preference surrogate model is adopted to quickly predict task accuracy rates for any sampled pair without a costly evaluation.
To demonstrate the effectiveness of the BLMOL framework, extensive experiments are conducted on three real-world graph datasets (ENZYMES, PROTEINS, and DHFR) to simultaneously handle multiple high-profile graph tasks, including GC, NC, and LP. Experimental results demonstrate that the MTGNN model obtained by the BLMOL outperforms state-of-the-art hand-crafted models. In addition, BLMOL can provide multiple models that match various user's task preferences. In summary, the key achievements of this paper are three-fold.

\begin{itemize}
  \item To the best of our knowledge, this paper is the first to explicitly discuss the BL-MOP in machine learning. Because an UL solution is associated with multiple high-dimensional LL Pareto optimal weights, this problem is computationally expensive.
  \item To address the aforementioned computational challenges, we introduce a simple but effective surrogate-assisted BLMOL framework by coupling the LL decision-making into the UL search and customizing the inexpensive preference surrogate model.
  \item To improve the limited representational capabilities of handcrafted GNNs and efficiently deal with possible task conflicts, we apply the proposed BLMOL to an innovative case study on multi-task graph neural topology search for the first time.
\end{itemize}

The remainder of this paper is organized as follows. Section II introduces the background and related works on BL-MOPs. The motivation for our studies is further highlighted. We present details of the proposed bi-level multi-objective learning framework in Section III. In Section IV, we apply the proposed method to solve the multi-task graph neural topology search problem. A series of experimental results are presented in Section V along with the ablation study. In section VI, conclusions and potential directions for future research are concluded.

\section{Background and Related Works}
In this section, we first introduce the background knowledge on BLMOPs in machine learning. Then, existing methods for BL-MOP are given to emphasize our motivation. As each sub-problem in BL-MOP involves multiple objectives, multi-objective optimization (MOO) in machine learning is also discussed. Since multi-task graph neural topology search is employed as a case study, related works on multi-task graph representation learning (MTGRL) and graph NAS (GNAS) are briefly outlined.

\subsection{Background Knowledge}

In general, given a target dataset $\mathcal{D} = \{\mathcal{D}_{tra}, \mathcal{D}_{val}, \mathcal{D}_{tet}\}$ drawn from a data sampling distribution $P_\mathcal{D}$, a simple unconstrained BL-MOPs in machine learning can be formulated as \cite{9766417}:
\begin{equation}\label{eq1}
\begin{split}
\underset{\bm{\alpha} \in \mathcal{A},\bm{w} \in \mathcal{W}}{\operatorname{minimize}} & \ \bm{\mathcal{L}}(\bm{\alpha}, \bm{w})\\
s.t. &\ \bm{w}\in \underset{\bm{w} \in \mathcal{W}}{\operatorname{argmin}} \ \bm{\mathcal{F}}(\bm{\alpha}, \bm{w})\\
\end{split},
\end{equation}
where $\bm{\mathcal{L}}(\bm{\alpha}, \bm{w})
=\left(\mathcal{L}_{1}(\bm{\alpha}, \bm{w}), \ldots, \mathcal{L}_{m}(\bm{\alpha}, \bm{w})\right)^\mathsf{T}$ is the UL vector-valued function with $m$ objectives on the validation set $\mathcal{D}_{val}:=\{(\bm{x^{(i)}},\bm{y^{(i)}}), \bm{x^{(i)}}\in \mathcal{X}, \bm{y^{(i)}}\in \mathcal{Y}\}_{i=1}^{N_V}$. $\bm{\mathcal{F}}(\bm{\alpha}, \bm{w})=\left(\mathcal{F}_{1}(\bm{\alpha}, \bm{w}),\ldots, \mathcal{F}_{m'}(\bm{\alpha}, \bm{w})\right)^\mathsf{T}$ is the LL vector-valued function with $m'$ objectives on the training set $\mathcal{D}_{tra}:=\{(\bm{x^{(i)}},\bm{y^{(i)}}), \bm{x^{(i)}}\in \mathcal{X}, \bm{y^{(i)}}\in \mathcal{Y}\}_{i=1}^{N_T}$. $\bm{\alpha}$ and $\bm{w}$ are the UL variable and learnable LL weight, respectively. UL variables could represent architectures, hyperparameters, meta-parameters and features, etc.

Since LL-MOP involves multiple conflicting objectives, there are multiple LL Pareto weights. Let $\bm{W^*}(\bm{\alpha})$ be a LL Pareto weight set for a given UL solution $\bm{\alpha}$:
\begin{equation}\label{eq1-1}
\bm{W^*}(\bm{\alpha})=\underset{\bm{w} \in \mathcal{W}}{\operatorname{argmin}}\ \bm{\mathcal{F}}(\bm{\alpha}, \bm{w}),
\end{equation}
then (\ref{eq1}) can be rewritten as follows:
\begin{equation}\label{eq1-3}
\begin{split}
 \underset{\bm{\alpha} \in \mathcal{A},\bm{w} \in \mathcal{W}}{\operatorname{minimize}} &\ \bm{\mathcal{L}}(\bm{\alpha}, \bm{w})\\
 s.t.&\ \bm{w}\in \bm{W^*}(\bm{\alpha})\\
\end{split}.
\end{equation}

In (\ref{eq1-3}), the BL-MOP searches the overall solution $(\bm{\alpha}, \bm{w})$ to minimize multiple UL objectives. \textcolor{red}{For a feasible overall solution $(\bm{\alpha}, \bm{w})$, the hierarchical scheme for the BL-MOP is that LL weight $\bm{w}$ must be one of the LL Pareto weight set $\bm{W^*}(\bm{\alpha})$ with respect to $\bm{\alpha}$.} Therefore, LL weight $\bm{w}$ needs to be selected from the LL Pareto weight set $\bm{W^*}(\bm{\alpha})$ so that the overall solutions $(\bm{\alpha}, \bm{w})$ are optimal at the UL. In this paper, we refer to it as the decision process of the LL-MOP.

\subsection{BL-MOPs}

 Due to the complex interaction mechanism between the UL and LL and the high computational cost, there are limited works on the BL-MOPs\cite{7942105}. Given the difficulty of BL-MOPs, classical methods often consider well-behaved mathematical problems. Eichfelder \cite{eichfelder2010multiobjective} proposed a nested approach that uses numerical optimization techniques and adaptive exhaustive search methods to solve LL and UL sub-problems, respectively. Shi \textit{et al.} \cite{shi2001model} presented an interaction algorithm for bi-level multi-objective decision-making with multiple interconnected decision-makers. Although computationally simple and efficient, these methods require some strong mathematical assumptions, which make them incapable of being directly applicable to many practical scenarios.

In recent years, some evolutionary methods have been proposed to solve BL-MOPs. Deb \textit{et al.} \cite{6793197} presented a hybrid evolutionary-local-search approach for solving the proposed challenging BL-MOP test problems. To save computational resources, Sinha \cite{sinha2011bilevel} proposed a progressively interactive evolutionary multi-objective optimization algorithm towards a preferred solution, where preference information comes from the UL decision maker. Ruuska \cite{ruuska2012constructing} presented a novel evolutionary bi-level multiobjective optimization algorithm with a partial order. More recently, Said \textit{et al.} \cite{9154441} proposed a multi-population transfer method for combinatorial BL-MOPs that considered a balance between diversity and convergence at the UL and selected individuals with the maximal contribution in terms of the indicator from the LL Pareto front. A knowledge-based dynamic variable decomposition co-evolutionary algorithm \cite{9721406} was proposed to solve the inseparable/separable BL-MOPs. The method applied an interaction matrix to group variables that are optimized in a cooperative manner. Wang \textit{et al.} \cite{doi:10.1080/09540091.2022.2077312} designed a double-population LL search strategy, where the first population maintains the convergence and diversity in the LL objective space, and the second population remains the non-dominance at both the UL and LL. Furthermore, Deb \textit{et al.} \cite{9766417} introduced a simple and efficient bi-level multi-objective optimization algorithm to minimize the deviation in the expected outcome of the UL due to the independent decision-making process for LL. The aforementioned evolutionary methods need to maintain multiple LL solutions during the optimization process. Despite good performance on low-dimensional BL-MOPs, these methods are not directly applicable to high-dimensional BL-MOPs in machine learning due to the expensive computational cost of maintaining multiple high-dimensional LL weights.

In the machine learning community, researchers have proposed various bi-level multi-objective algorithms in the context of different applications, such as multi-objective NAS\cite{lu2020nsganetv2,LOUATI202144,9508774,9430615}, multi-task learning (MTL)\cite{Raychaudhuri_2022_CVPR}, and multi-objective meta learning\cite{ye2021multi}. However, these studies mainly focus on scenarios with multiple UL objectives and one LL objective. More complex problems in which both the UL and LL are multi-objective are rarely studied.
A common intuition is to build a proxy LL objective that optimizes a weighted linear combination of all task losses at the LL. However, this method needs to manually adjust the loss's weights in the proxy LL objective, which could be time-consuming\cite{ye2021multi,9766417}. In this paper, we propose a novel BLMOL framework to avoid the above issues.

\subsection{MOO}
In a machine learning system, MOO seeks to find a set of Pareto solutions that represent different trade-offs between possibly conflicting objectives \cite{navon2021learning}. Existing massive works on the MOO in machine learning can be mainly divided into 1) population-based methods and 2) gradient-based methods.

Population-based MOO methods \cite{sharma2022comprehensive} mainly include dominance-based\cite{zitzler2001spea2,deb2002fast}, decomposition-based\cite{4358754,9819828}, indicator-based\cite{ishibuchi2010indicator,10.1162/EVCO_a_00009}, hybrid-based\cite{6600851,6964796}, and model-based methods\cite{8580560,9082904,9817389}. Due to their easy scalability and gradient-free properties, these population-based methods are widely used in various machine learning problems, such as neuroevolution\cite{Neuroevolutionary,https://doi.org/10.48550/arxiv.2106.07611}, NAS\cite{elsken2018efficient,lu2020nsganetv2,hu2021accelerating,https://doi.org/10.48550/arxiv.2207.05321}, feature selection\cite{7339682,9641743}, reinforcement learning\cite{pmlr-v119-xu20h}, federated learning\cite{8744465,9762229}, MTL\cite{9504721}, and fairness learning\cite{QingquanFair2021}. In addition, many surrogate-assisted multi-objective evolutionary algorithms have been proposed to solve those machine learning systems with expensive optimization objectives \cite{8456559,8744404,liu2022survey}.

In recent years, many gradient-based MOO methods for deep learning have been proposed. Due to the inevitable task conflict when learning a model for multiple machine learning tasks simultaneously, most of these methods take MTL as an example\cite{sener2018multi,lin2019pareto,ma2020efficient,mahapatra2020multi,lin2020controllable,navon2021learning,ruchte2021scalable}. Sener \textit{et al.} \cite{sener2018multi} initially  employed MGDA\cite{DESIDERI2012313} to train a deep MTL model to obtain a Pareto stationary solution. Many follow-up preference-based methods are proposed to produce a set of well-distributed Pareto solutions by introducing preference vectors \cite{lin2019pareto,mahapatra2020multi}. For example, the exact Pareto optimal (EPO) search aims to find an exact Pareto optimal solution for a given preference vector by combining controlled ascent and gradient descent\cite{mahapatra2020multi}. Furthermore, as the most straightforward MOO method, linear scalarization (LS) with preference vectors as weights is also widely used in deep learning\cite{navon2021learning}.

\subsection{MTGRL and GNAS}
Most of the work on MTGRL focused on various applications on graphs. A unified MTGNN is proposed by Avelar \textit{et al.} \cite{avelar2019multitask}, which learns centrality measures for multiple graphs. Holtz \textit{et al.} \cite{holtz2019multi} presented a multi-task supervised learning paradigm for bi-level labeling. Li \textit{et al.} \cite{li2019syntax} developed a syntax-aware MTGNN for the biomedical relation extraction and classification task. An MTGNN was proposed by Montanari \textit{et al.} \cite{montanari2019modeling} to model physical-chemical ADMET endpoints. Wang \textit{et al.} \cite{wang2020multi} introduced a multi-task network embedding method, where one task maintains high-order proximity across paired nodes throughout the network and the other task aims to maintain low-order proximity in the single-hop region of each node. Xie \textit{et al.} \cite{xie2020multi1} presented an MTGRL framework for enhanced graph classification, which exploits the available knowledge of the node classification to better improve the performance of the graph classification. Then they further developed \cite{xie2020multi2} MTGRL to solve the node classification and link prediction. Buffell \textit{et al.} \cite{buffelli2021graph} proposed a meta-learning framework to obtain node embeddings suitable for multiple tasks, simultaneously. However, these handcrafted approaches for specific tasks rely on expert knowledge and ignore possible conflicts across tasks.

GNAS aims to automatically discover the best neural architecture configuration for different graph-related tasks without manual design. The challenges are typically divided into designing the search space and the search strategy\cite{wang2021automated}. According to the construction principles of graph neural modules, the search space can be divided into four categories: micro search space, macro search space, pooling methods, and hyperparameters. Numerous studies \cite{9458743,10.1145/3485447.3512185} have demonstrated that automating the design of the best aggregation operation and network topology can improve model capacity. Search strategies include reinforcement learning, differentiable methods, evolutionary algorithms, and hybrid methods. Existing works primarily designed the search space and search strategy to learn a graph representation model customized for a given single task. However, automatic learning methods for multiple tasks on graphs have not been noticed.

In summary, MTGNN topology search is first presented to improve the limited representation ability of handcrafted GNNs through cross-task knowledge sharing. The problem naturally is viewed as a typical BL-MOP in machine learning. We use the proposed BLMOL framework to find a set of MTGNN topologies and their weights, representing different trade-offs among different tasks at UL and LL, respectively.

\section{BLMOL Framework}
This section introduces the algorithmic details of the BLMOL framework. Previously, we have described the form of BL-MOP in machine learning. Here, we further discuss how to formally couple the LL decision-making process with the UL optimization process. Then we describe a constructed preference surrogate model for replacing the expensive evaluation process at the UL. Finally, the whole framework of BLMOL is presented.

\subsection{Problem Transformation}
The decision-making process of the LL-MOP aims to select a LL weight $\bm{w}$ from the LL Pareto weight set $\bm{W^*}(\bm{\alpha})$ for an UL solution $\bm{\alpha}$ so that overall solutions $(\bm{\alpha}, \bm{w})$ is optimal at the UL. A LL Pareto weight set $\bm{W^*}(\bm{\alpha})$ needs to be found for any feasible UL solution $\bm{\alpha}$, which leads to a high computational cost for the BL-MOP. Therefore, we do not recommend solving (\ref{eq1-3}) directly. In this paper, we couple the decision-making process of the LL-MOP to the search process of the UL-MOP by introducing the LL preference vector $\bm{r}=(r_1,r_2,\ldots,r_{m'})$. Then (\ref{eq1-3}) can be transformed as follows:
\begin{equation}\label{eq4-1}
\begin{split}
\underset{(\bm{\alpha},\bm{r}) \in \mathcal{A} \times \mathcal{R}}{\operatorname{minimize}} &\ \bm{\mathcal{L}}(\bm{\alpha}, \bm{w^*}(\bm{\alpha},{\bm{r}}))\\
s.t.&\ \bm{w^*}(\bm{\alpha},{\bm{r}})\in\underset{\bm{w} \in \mathcal{W}}{\operatorname{argmin}}\ \bm{\mathcal{F}}(\bm{\alpha}, \bm{w})\\
\end{split}.
\end{equation}

In (\ref{eq4-1}), the BL-MOP searches the UL variables $\bm{\alpha}$ and LL preference vectors $\bm{r}$ to minimize UL objectives. While the LL-MOP is transformed into a preference MOO problem. \textcolor{red}{Comparing (\ref{eq1-3}) and (\ref{eq4-1}), the purpose of the LL-MOP is the same, that is, to provide a Pareto weight $\bm{w}$ with respect to UL variable $\bm{\alpha}$. Different from (\ref{eq1-3}), by introducing preference $\bm{r}$, the LL-MOP in (\ref{eq4-1}) aims to obtain an exact Pareto weight with a preference vector, which is defined as follows \cite{mahapatra2020multi}:}

Given an UL variable $\bm{\alpha}$, a LL weight is said to be an exact LL Pareto weight $\bm{w}^*$ with respect to a LL preference vector $\bm{r}$ 
if and only if two conditions are met:

1) $\bm{w^*}$ is a LL Pareto weight; 

2) $r_1\mathcal{F}_{1}(\bm{\alpha},\bm{w^*})=r_2\mathcal{F}_{2}(\bm{\alpha},\bm{w^*})=\ldots=r_{m'}\mathcal{F}_{m'}(\bm{\alpha},\bm{w^*})$.

\begin{figure}
\centering
\includegraphics[width=0.3\textwidth]{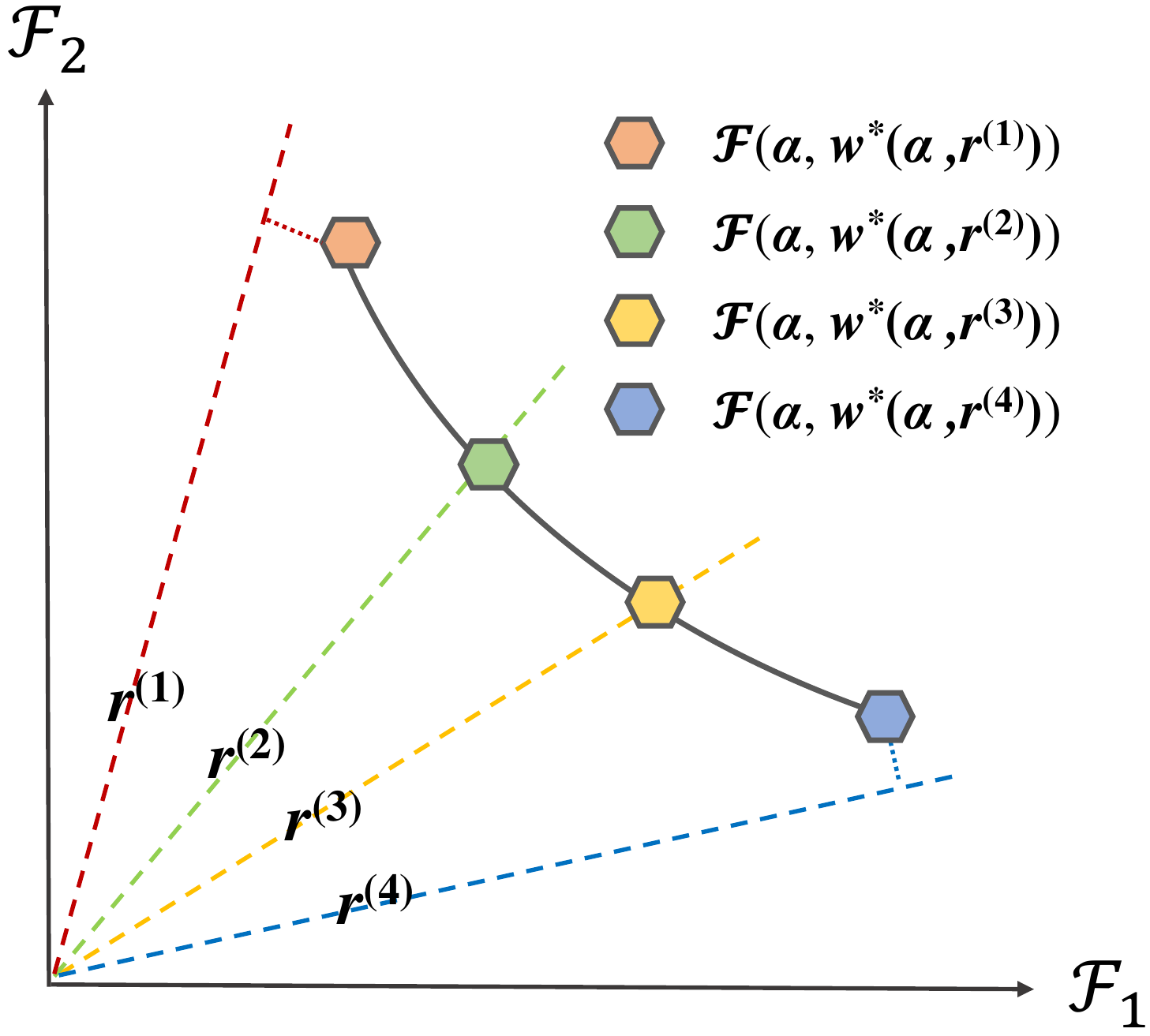}
\caption{\textcolor{red}{An illustrative example of the exact LL Pareto weight. When $\bm{r}$ and the LL Pareto front are disjoint, the LL-MOP aims to find a LL Pareto weight that is closest to $\bm{r}$, such as $\bm{r}^{(1)}$ and $\bm{r}^{(4)}$.}} \label{LLEPO}
\end{figure}

\textcolor{red}{As illustrated in Fig \ref{LLEPO}, for any exact LL Pareto weight $\bm{w^*}$, $\bm{\mathcal{F}(\bm{\alpha},\bm{w}^*)}$ is the intersection of the preference vector $\bm{r}$ and the LL Pareto front. When the given LL preference vector $\bm{r}$ and the LL Pareto front are disjoint, the LL-MOP aims to find a LL Pareto weight that is closest to the preference vector $\bm{r}$\cite{mahapatra2020multi}.} Compared to finding a LL Pareto weight set, our method only needs to find an exact LL Pareto weight, which significantly saves computational resources.

\subsection{Preference Surrogate Model Construction}
Since the evaluation process of the UL-MOP needs to complete a LL optimization and the LL-MOP often involves complex model training, the calculation of the UL-MOP is costly. To alleviate the evaluation cost, we adopt a preference surrogate model to fit the functional relationship between UL variables, LL preference vectors, and multiple UL objective functions. The surrogate model needs to meet the following three requirements: (1) Low computational cost; (2) High-order correlation across the predicted and the true UL objective function; (3) Given the UL variable and the LL preference vector, the predicted value satisfies the preference optimality of the LL-MOP.

\begin{figure}
\centering
\includegraphics[width=0.4\textwidth]{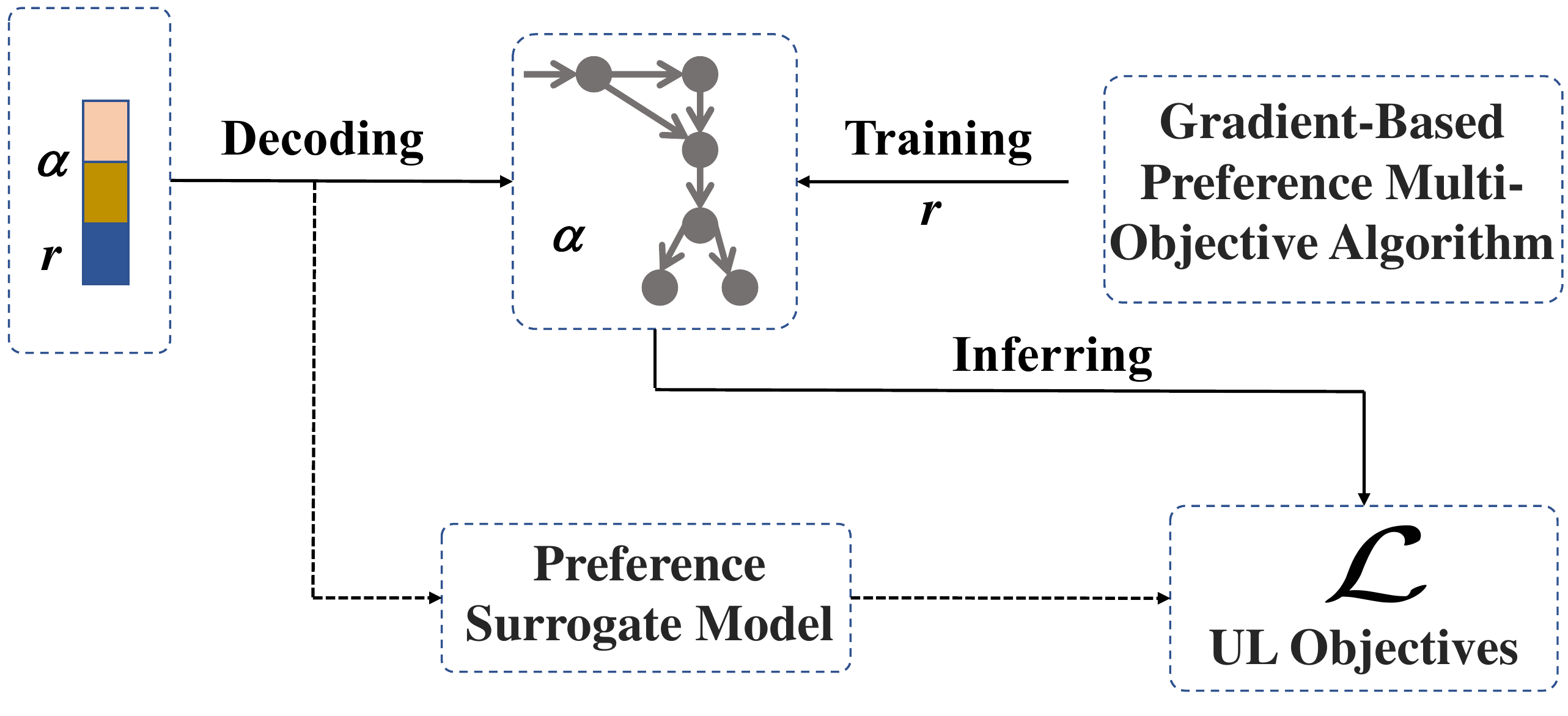}
\caption{An illustrative example of the construction process of the preference surrogate model on NAS.} \label{fig3}
\end{figure}

The whole construction process of the preference surrogate model is depicted in Algorithm \ref{alg1} and Fig. \ref{fig3}. Firstly, we sample a set of UL variables $\bm{\alpha}$ and LL preference vectors $\bm{r}$ from the joint space of $\mathcal{A}$ and $\mathcal{R}$. For each pair $(\bm{\alpha}^{(j)},\bm{r}^{(j)})$, LL weight $\bm{w}^{(j)}$ is trained to minimize multiple LL objectives using a GPMOA over multiple epochs, such as LS\cite{navon2021learning} and EPO\cite{mahapatra2020multi}. The descent direction of these methods is essentially a convex combination of gradients. The model weights $\bm{w}$ can be updated by

\begin{equation}
\label{LLMOP}
\bm{w} = \bm{w} - \xi \bm{\mu}^T\nabla_{\bm{w}} \bm{\mathcal{F}}(\bm{\alpha}, \bm{w}),
\end{equation}
where $\xi$ and $\bm{\mu}=(\mu_1,\cdots,\mu_{m'})^\mathsf{T}$ are the learning rate and combination coefficient, respectively. In LS, the preference vector $\bm{r}$ is regarded as the combination coefficient, i.e. $\bm{r} = \bm{\mu}$ \cite{navon2021learning}. Even with some theoretical limitations, LS is a fast method that tends to work well in practice\cite{navon2021learning}. In EPO, the combination coefficient is adaptively calculated by solving a linear programming problem, which seeks to find a direction that makes $\bm{w}$ have better convergence or uniformity, or both\cite{mahapatra2020multi}.

After the model is trained, we can easily calculate the UL objective values. Finally, multiple preference surrogate models for each UL objective $\mathcal{L}_i$ are fitted using the pair of data consisting of $(\bm{\alpha}^{(j)},\bm{r}^{(j)}$ and $\mathcal{L}_i^{{(j)}})$. Commonly used cheap surrogate models include support vector regression (SVR), random forest (RF), AdaBoost, bagging, Gaussian process (GP), and multi-layer perceptron (MLP)\cite{jin2021data}. Because no single surrogate model outperforms other models on all datasets or objectives, we construct multiple different types of surrogate models and then select the best model for each UL objective via cross-validation in this paper. 

\begin{algorithm}[t]
 \caption{Preference Surrogate Model (PSM)} 
 \label{alg1}
 \begin{algorithmic}[1]
  \REQUIRE $\mathcal{M}$: Machine learning model; $\mathcal{D}_{train}$: Training set; $\mathcal{D}_{val}$: Validation set; $m$: Number of UL objectives; $N_s$: Number of samples.
  \ENSURE $SM_{i},i=1,...,m$: Preference surrogate model set.
  \STATE $i \leftarrow 0$ // Initialize an counter;
  \STATE  $(\bm{\alpha}^{(j)}, \bm{r}^{(j)}), j=1, ..., N_s\leftarrow$ Uniformly sample $N_s$ UL variables and LL preference vectors from $\mathcal{A} \times \mathcal{R}$;
  \STATE $\bm{w}^{(j)}, j=1,...,N_s\leftarrow$ Train $N_s$ machine learning models on the training set with different $(\bm{\alpha}^{(j)}, \bm{r}^{(j)})$ by (\ref{LLMOP}) over multiple epochs;
  \STATE $\mathcal{L}^{{(j)}}=(\mathcal{L}_1^{{(j)}}, ..., \mathcal{L}_m^{{(j)}}),j=1,...,N_s\leftarrow$ Calculate the UL objective values of $(\bm{\alpha}^{(j)}, \bm{r}^{(j)})$ on the validation set;
  \WHILE{$i < m$}
  \STATE $(\bm{\alpha}^{(j)}, \bm{r}^{(j)},\mathcal{L}_i^{{(j)}}) \leftarrow$ Obtain the training data pair;
  \STATE $(SM_i^1,...,SM_i^p) \leftarrow$ Fit $p$ preference surrogate models by $(\bm{\alpha}^{(j)}, \bm{r}^{(j)},\mathcal{L}_i^{{(j)}})$;
  \STATE $SM_i \leftarrow$ Select the best preference surrogate model from $(SM_i^1,...,SM_i^p)$;
  \STATE $i\rightarrow{i+1}$;
  \ENDWHILE
 \end{algorithmic}
\end{algorithm}

\subsection{Framework of BLMOL}
\begin{algorithm}[t]
 \caption{Framework of BLMOL} 
 \label{alg2}
 \begin{algorithmic}[1]
  \REQUIRE $\mathcal{M}$: Machine learning model; $\mathcal{D}_{train}$: Training set; $\mathcal{D}_{val}$: Validation set; $m$: Number of UL objectives; $N_s$: Number of samples; $T$: Number of iterations; $N_p$: Population size; $MOO$: Multi-objective optimizer.
  \ENSURE $P$: Final archive.
  \STATE $(SM_{1},...,SM_{m})\leftarrow PSM(\mathcal{M}, m, N_s, \mathcal{D}_{train}, \mathcal{D}_{val})$;  // Algorithm \ref{alg1}
  \STATE $t \leftarrow 0$ // Initialize an counter;
  \STATE $P \leftarrow \emptyset $ // Initialize a population;
  \WHILE{$j < N_p$}
  \STATE $(\bm{\alpha}^{(j)}, \bm{r}^{(j)}) \leftarrow$ Sample$(\mathcal{A} \times \mathcal{R})$;
  \STATE $(\mathcal{L}_1^{{(j)}}, ..., \mathcal{L}_m^{{(j)}})\leftarrow SM_{1}(\bm{\alpha}^{(j)}, \bm{r}^{(j)}), ..., SM_{m}(\bm{\alpha}^{(j)}, \bm{r}^{(j)})$;
  \STATE $P \leftarrow P \cup (\bm{\alpha}^{(j)}, \bm{r}^{(j)})$;
  \STATE $j\rightarrow{j+1}$;
  \ENDWHILE
  \WHILE{$t < T$}
      \STATE $P \leftarrow MOO(P,(SM_{1},...,SM_{m}))$;
      \STATE $t\rightarrow{t+1}$;
  \ENDWHILE
  \STATE $P \leftarrow NDsort(P)$.
 \end{algorithmic}
\end{algorithm}

According to the characteristics of the BL-MOP, we employ a MOEA to minimize UL objectives. The basic process of BLMOL is presented in Algorithm \ref{alg2}. Firstly, the preference surrogate models are constructed for each UL objective (see Algorithm \ref{alg1}). Then we initialize a population by sampling a set of UL variables and LL preference vectors and calculating the value of each UL objective function using the preference surrogate models. Next, based on the encoding of the solution, the population is evolved by a MOEA, such as NSGAII\cite{deb2002fast} and MOEA/D\cite{4358754}. The MOEA maintains a population to approximate the Pareto front in the objective space through reproduction and natural selection. Finally, the non-dominated solution is selected from the population as the output.

BLMOL consists of two main steps which are preference surrogate model construction and UL multi-objective evolutionary search. Since the training cost of the surrogate model is much lower than that of the LL weights, the complexity of constructing the preference surrogate model is mainly determined by the LL weight training (Step 3 of Algorithm 1). Suppose EPO\cite{mahapatra2020multi} is embedded into BL-MOL to train the weight of each sample. EPO scales linearly with the dimension $(n)$ of the weight, whose time complexity is $O(m^{'2}n)$\cite{mahapatra2020multi}. $m^{'}$ is the number of LL objectives. Therefore, the time complexity of constructing the preference surrogate model is $O(N_sN_em^{'2}n)$, where $N_s$ and $N_e$ are the sample size and the epoch size of EPO, respectively. Assume that NSGAII\cite{deb2002fast} are embedded in BL-MOL to optimize the $m$ cheap UL surrogate objectives. The time complexity of NSGAII is $O(mN_p^2T)$, where $N_p$ and $T$ are the population size and the number of iterations, respectively. In deep learning, $n$ is much larger than other values $(m^{'}, m, N_s, N_e$, and $N_p)$. Therefore, the overall time complexity of BLMOL is $O(N_sN_em^{'2}n)$.

\section{MTGNTS}
In this section, we adopt the proposed BLMOL to solve the multi-task graph neural topology search problem. Preliminary knowledge and problem statement are given first. The customized search space and search strategy are then described.

\subsection{Preliminary Knowledge}
Graph-structured data is employed to model the complex relationships across entities in real-world physical systems, such as social networks\cite{wu2021pareto}, protein-protein interaction networks\cite{wang2004identifying}, and traffic networks\cite{strogatz2001exploring}. Let $\bm{G}=(\bm{V},\bm{E})$ be a graph-structured data with node features $\bm{X}_i \in \bm{R}^{d}$ for $i\in \bm{V}$. As the most popular tool for learning graph representations, GNNs capture graph embedding information through message passing with neighbors on the graph data, formulated as follows:
\begin{equation}\label{eqGNN}
    \mathbf{h}_i^{(k)}=\mathrm{ACT}^{k}\left( \mathbf{W}^{(k)} \cdot \mathrm{AGG}^{(k)}\left(\left\{ \mathbf{h}_{j}^{(k-1)}, \forall j \in \mathcal{N}(i)\right\}\right)\right),
\end{equation}
where $\mathbf{h}_i^{(k)}$, $\mathrm{ACT}^{k}$, $\mathbf{W}^{(k)}$, and $\mathrm{AGG}^{(k)}$ refer to the node representation of node $i$, the activation  operation, learnable weights, and the aggregation function at layer $k$, respectively. $\mathcal{N}(i)=\{j\in \bm{V}|(i,j)\in \bm{E} \}$ is the neighborhood of node $i$. For a more concise representation, we define $\mathrm{ACT}^{k}\left( \mathbf{W}^{(k)} \cdot \mathrm{AGG}^{(k)}\left( * \right)\right) = \mathrm{A}^{(k)}\left( * \right) $.

\begin{figure*}
	\centering
	\subfloat[Vanilla GNN]{
		\includegraphics[width=0.2\linewidth]{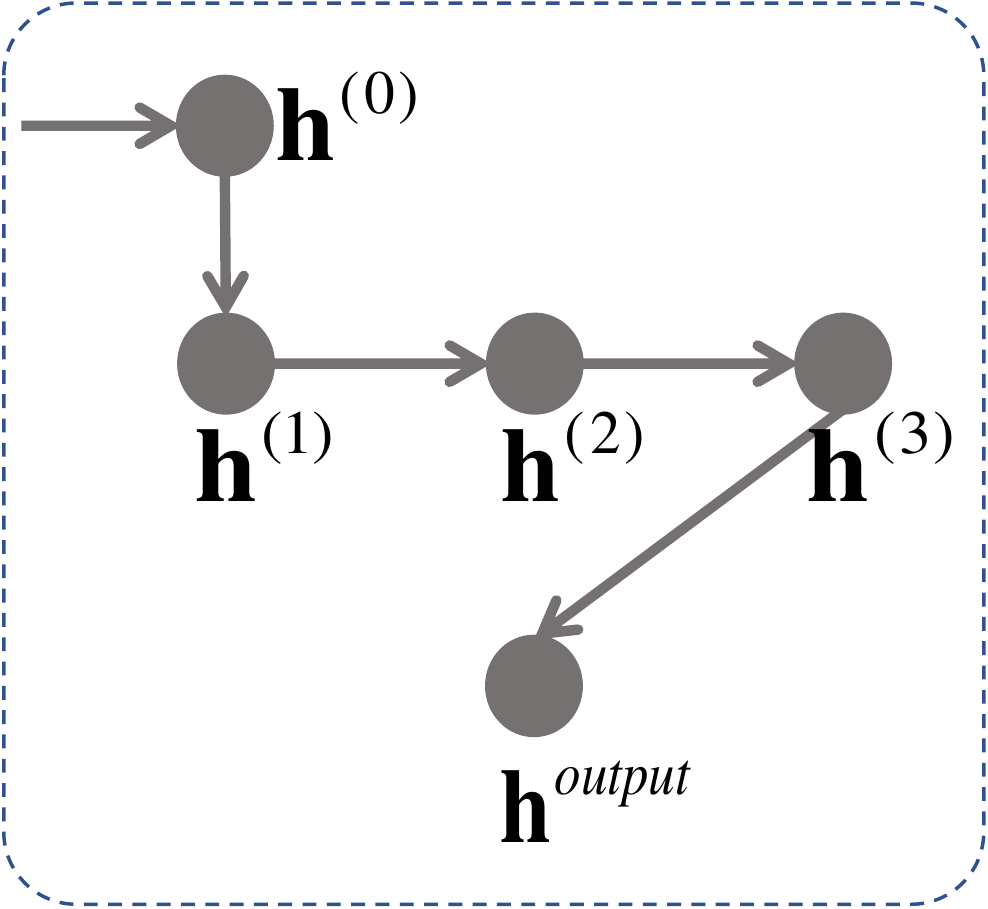}}
	\hfill
	\subfloat[ResGNN]{
		\includegraphics[width=0.2\linewidth]{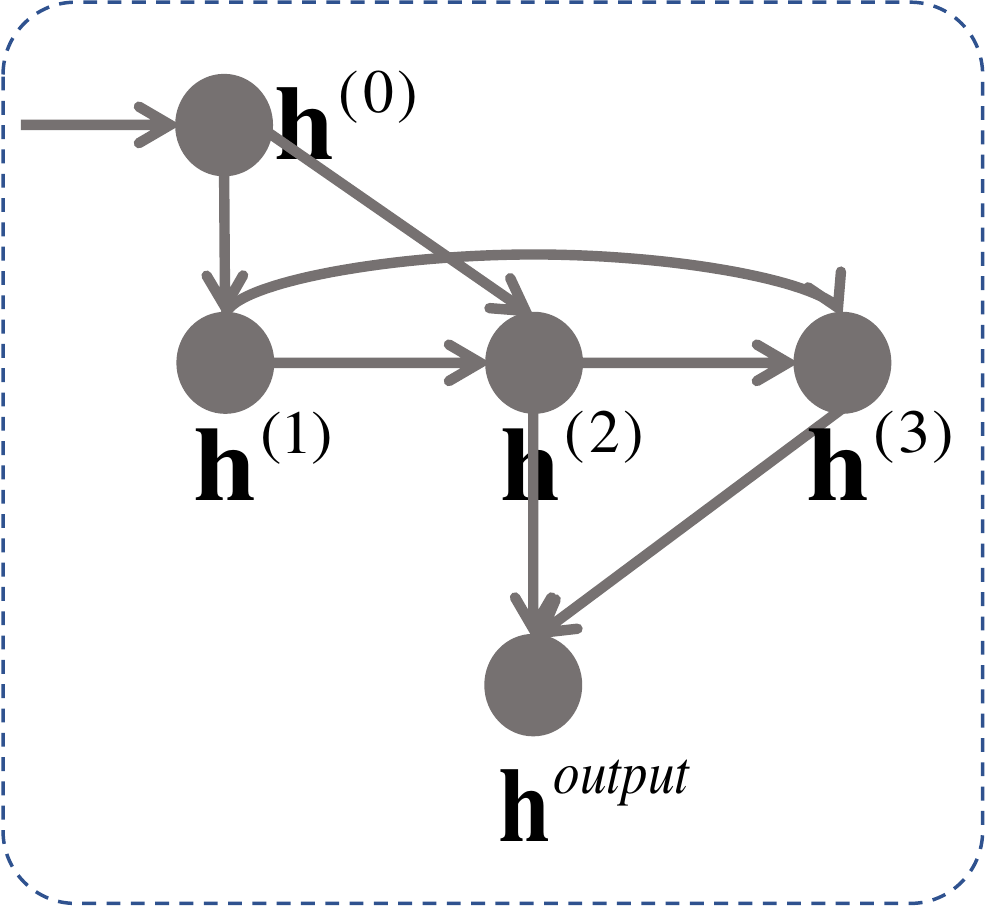}}
	\hfill
	\subfloat[JK-Net]{
		\includegraphics[width=0.2\linewidth]{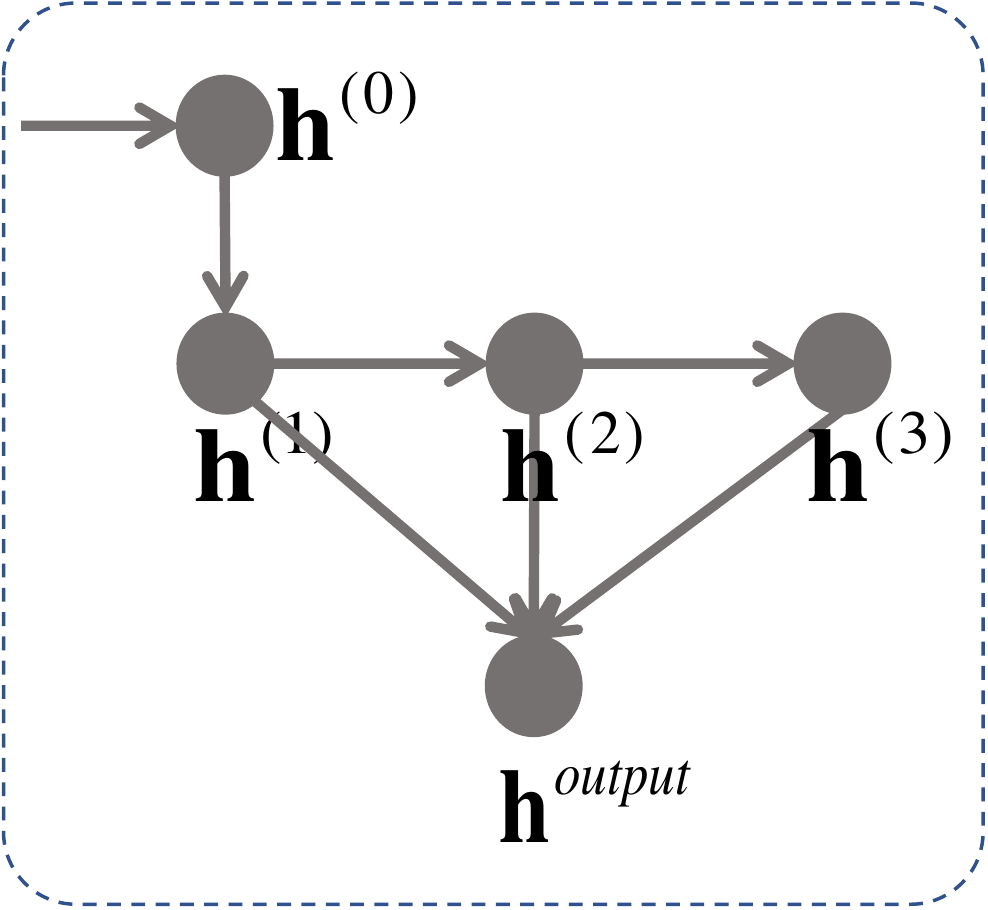}}
	\hfill
	\subfloat[F$^2$]{
		\includegraphics[width=0.2\linewidth]{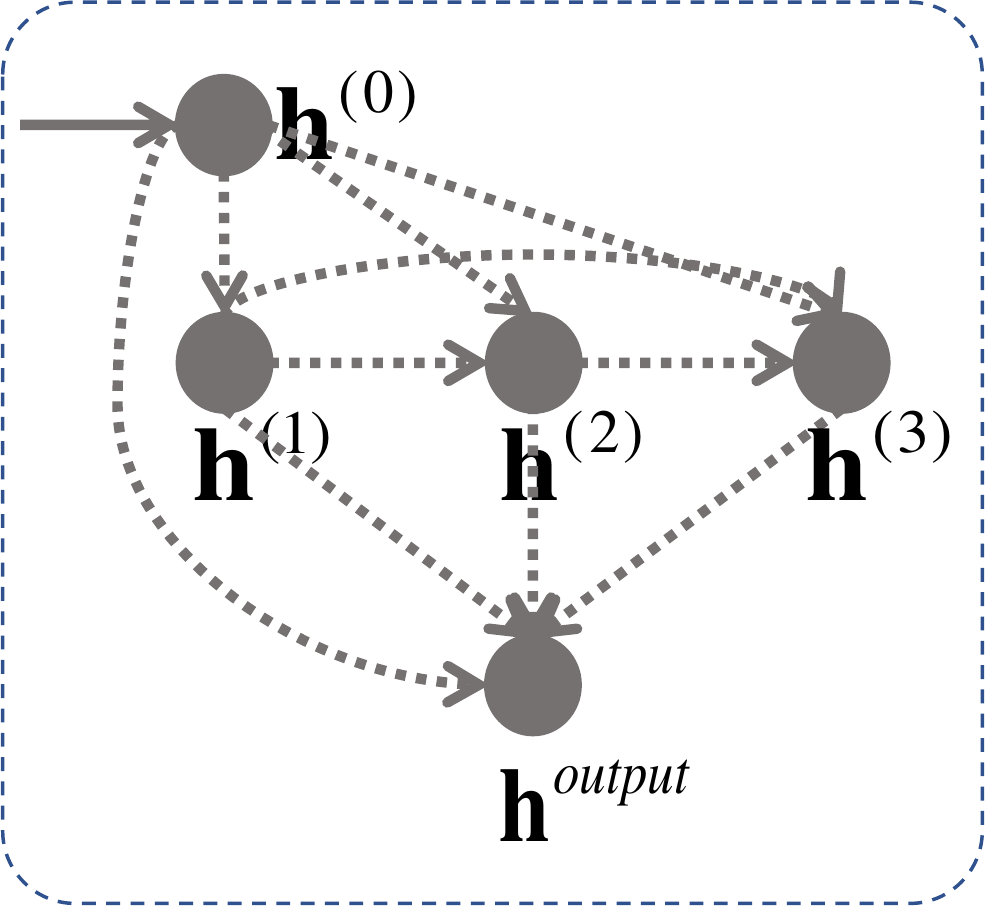}}
	\caption{Several common topologies of GNNs (Taking a three-layer backbone as an example).}
	\label{fig4}
\end{figure*}

Considering the GNN model with skip connections \cite{xu2018representation, chen2020simple,10.1145/3485447.3512185}, the node representation $\mathbf{h}^{(k)}$ at layer $k$ is updated as:

\begin{equation}\label{eqGNNlayer}
\begin{aligned}
    \mathbf{h}^{(k)} = \mathrm{A}^{(k)}  \left(\mathrm{F}^{(k)} \left( \mathrm{S}^{(k)}_0 \left( \mathbf{h}^{(0)}\right),\cdots,\mathrm{S}^{(k)}_{k-1}\left(\mathbf{h}^{(k-1)} \right)\right) \right),
\end{aligned}
\end{equation}
where $\mathrm{A}^{(k)}$ and $\mathrm{F}^{(k)}$ are aggregator operations and fusion operations at layer $k$, respectively. Fusion operations are employed to fuse graph features, such as MAX, CONCAT, and LSTM. Aggregation operations determine the neighborhood aggregation schema, such as GCN, GAT, GIN, and SAGE. $\mathrm{S}^{(k)}_0$ is the skip connection for the node representation ${\mathbf{h}}^{(0)}$ at layer $k$ including ZERO and IDENTITY, which mean that ${\mathbf{h}}^{(0)}$ is not selected and selected, respectively. Skip connections determine the topology of the neural network. Fig. \ref{fig4} shows the topology of several GNNs. Vanilla GNNs \cite{xu2018how} perform aggregation operations on a single path:
\begin{equation}\label{Vanilla GNNs}
    \mathbf{h}^{(k)} = \mathrm{A}^{(k)}  \left( \mathbf{h}^{(k-1)} \right).
\end{equation}
And ResGNN \cite{Li_2019_ICCV} and JK-Net\cite{xu2018representation} design different skip connections by introducing residual connections, as shown in \ref{ResGCN} and \ref{JK-Net} respectively.
\begin{equation}\label{ResGCN}
    \mathbf{h}^{(k)} = \mathrm{A}^{(k)}  \left( \mathbf{h}^{(k-1)} + \mathbf{h}^{(k)} \right),
\end{equation}
\begin{equation}\label{JK-Net}
\left\{\begin{matrix}
 \mathbf{h}^{(k)} = \mathrm{A}^{(k)}  \left( \mathbf{h}^{(k-1)} \right)\\\mathbf{h}^{output} =  \mathrm{F}^{(k)} \left( \mathbf{h}^{(1)},\cdots,\mathbf{h}^{(k-1)} \right).
\end{matrix}\right.
\end{equation}

Further, Wei \textit{et al.} \cite{10.1145/3485447.3512185} proposed a unified topology framework F$^2$. And a GNAS method is presented to search for optimal topological architectures on F$^2$. In addition, for graph-level tasks, the node representation from the entire graph $\bm{G}$ is aggregated by a readout operation $\mathrm{R}$:
\begin{equation}\label{eqGT}
    \mathbf{h}_{\bm{G}}=\mathrm{R}\left(\left\{ \mathbf{h}_{j}^{(K)}, \forall j \in \bm{V} \right\}\right).
\end{equation}

\subsection{Problem Statement}
In a graph neural topology search problem for multiple machine learning tasks, we anticipate obtaining multiple Pareto topologies and their Pareto weights to satisfy users' needs for different task trade-offs. MTGNTS can be formalized in the following manner:
\begin{equation}\label{eq7}
\begin{split}
\underset{(\bm{\alpha},\bm{r}) \in \mathcal{A} \times \mathcal{R}}{\operatorname{maximize}} &\ \bm{Acc}(\bm{\alpha}, \bm{w^*}(\bm{\alpha},{\bm{r}}))\\
s.t.&\ \bm{w^*}(\bm{\alpha},{\bm{r}})\in\underset{\bm{w} \in \mathcal{W}}{\operatorname{argmin}}\ \bm{\mathcal{F}}(\bm{\alpha}, \bm{w})\\
\end{split},
\end{equation}
where $\bm{Acc}(\cdot)=\left({Acc}_{1}(\cdot), \ldots, {Acc}_{m}(\cdot)\right)^\mathsf{T}$ 
and $\bm{\mathcal{F}}(\cdot)=\left(\mathcal{F}_{1}(\cdot)),\ldots, \mathcal{F}_{m}(\cdot)\right)^\mathsf{T}$ represent the evaluation metrics and loss functions for $m$ graph tasks, respectively.
The UL-MOP searches the topology $\bm{\alpha}$ of the MTGNN and the LL preference vector $\bm{r}$ to maximize the evaluation metrics of the graph task on the validation set $\mathcal{D}_{val}$. For a given $(\bm{\alpha}, \bm{r})$, the weight $\bm{w}$ of the MTGNN is trained on the training set $\mathcal{D}_{tra}$ to minimize multiple task losses. This paper focuses on three important tasks in graph machine learning, including GC, NC, and LP.

GC employs graph-level features $\mathbf{h}_{\bm{G}}$ and labels $y_{\bm{G}}$ to predict labels for the entire graph $\bm{G}\in \mathcal{G}=\{\bm{G}^1,...,\bm{G}^n\}$. A squared-error loss can be employed:
\begin{equation}\label{GC}
\mathcal{F}_{1} = \sum_{\bm{G}\in \mathcal{G}}{||f(\mathbf{h}_{\bm{G}})-y_{\bm{G}}||_2^2}.
\end{equation}

NC aims to predict the labels of unknown nodes in a graph given a training set $\mathcal{V}_{tra}$ with node features $\mathbf{z}_u$ and labels $\mathbf{y}_u$. We can use a softmax classification function and negative log-likelihood loss: 

\begin{equation}\label{NC}
\mathcal{F}_{2} = \sum_{u\in \mathcal{V}_{tra}}{-log(softmax(\mathbf{z}_u,\mathbf{y}_u))}.
\end{equation}

LP predicts whether there is an unobserved true connection across two nodes in a graph, which can be transformed into a binary classification problem. The edges in the graph can be regarded as positive examples, and some non-existing edges can be viewed as negative examples. And they constitute a set of training data sets $\mathcal{D}_{tra}$. Then we can evaluate the model using any binary classification metric such as AUC. Binary cross-entropy loss can be used as the loss function:
\begin{equation}\label{LP}
\begin{gathered}
\mathcal{F}_{3}=-\sum_{(u,v) \in \mathcal{D}_{tra}}
(y_{(u,v)} log (\hat{y}_{(u,v)})+ \\
(1-y_{(u,v)}) log(1-\hat{y}_{(u,v)}))
\end{gathered}.
\end{equation}

\begin{figure}
\centering
\includegraphics[width=0.45\textwidth]{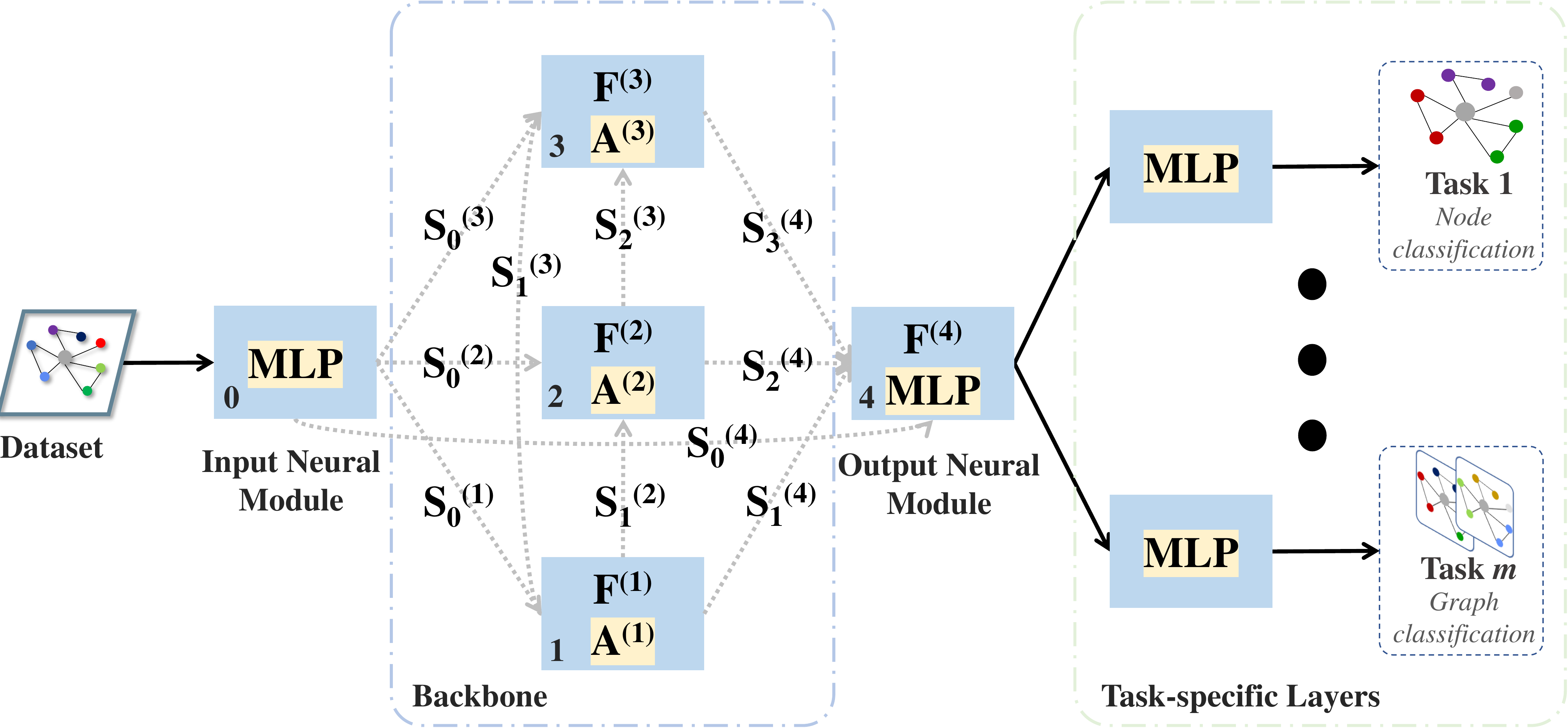}
\caption{An illustration of the MTGNN model for a three-layer backbone.} \label{fig5}
\end{figure}

\begin{table}
\centering
\caption{The search space in this paper.}\label{Search-Space}
\resizebox{\linewidth}{!}{
\begin{tabular}{c|c}
\toprule 
Operators & Value (Encoding) \\
\hline
Skip connections $\mathrm{S}$ & ZERO(0), IDENTITY(1)  \\ 
Fusion operations $\mathrm{F}$ & MEAN(0), CONCAT(1), LSTM(2), MAX(3), SUM(4), ATT(5)  \\ 
\bottomrule
\end{tabular}
}
\end{table}

\subsection{Search Space and Search Strategy}
Based on the unified topology framework F$^2$, we represent the MTGNN model as a supernet, which is a directed acyclic graph (DAG), as depicted in Fig. \ref{fig5}. The supernet is constructed with an input neural module, $K$ backbones, an output neural module, and $m$ task-specific layers. The input neural module is a simple preprocessing operation, i.e., MLP in this paper. The $k$-th backbone contains $k$ skip connections $\{\mathrm{S}_{0}^{(k)},...,\mathrm{S}_{k-1}^{(k)}\}$, an aggregation operation $\mathrm{A}^{(k)}$, and a fusion operation $\mathrm{F}^{(k)}$. In the output neural module, after skip connections $\{\mathrm{S}_{0}^{(K+1)},...,\mathrm{S}_{K}^{(K+1)}\}$ and a fusion operation $\mathrm{F}^{(K+1)}$, a MLP is adopted as a post-processing operation to obtain the embedding vector. Then the embedding vector is fed into task-specific layers to handle multiple downstream tasks. Details of task-specific layers can be found in \textit{Experiment Setting} (Section V. A \textit{b)}).

The choice of backbone topology and operation is important for model capacity \cite{9458743,10.1145/3485447.3512185}. In this paper, we mainly explore how to automatically search for the backbone topology and the fusion operation.
Without loss of generality, a set of candidate skip connections $\mathrm{S}$ and fusion operations $\mathrm{F}$ are considered in the search space $\mathcal{A}$ as shown in Table \ref{Search-Space}. If the computing resources are sufficient, more operations can be extended in the search space, such as aggregation operations and activation functions. For a three-layer backbone, the number of skip connections is $\sum_{i=1}^{3+1}i=10$, and the number of fusion operations is $4$. Then the size of the search space is $2^{10} + 6^{4}=2320$.

\begin{figure}
\centering
\includegraphics[width=0.45\textwidth]{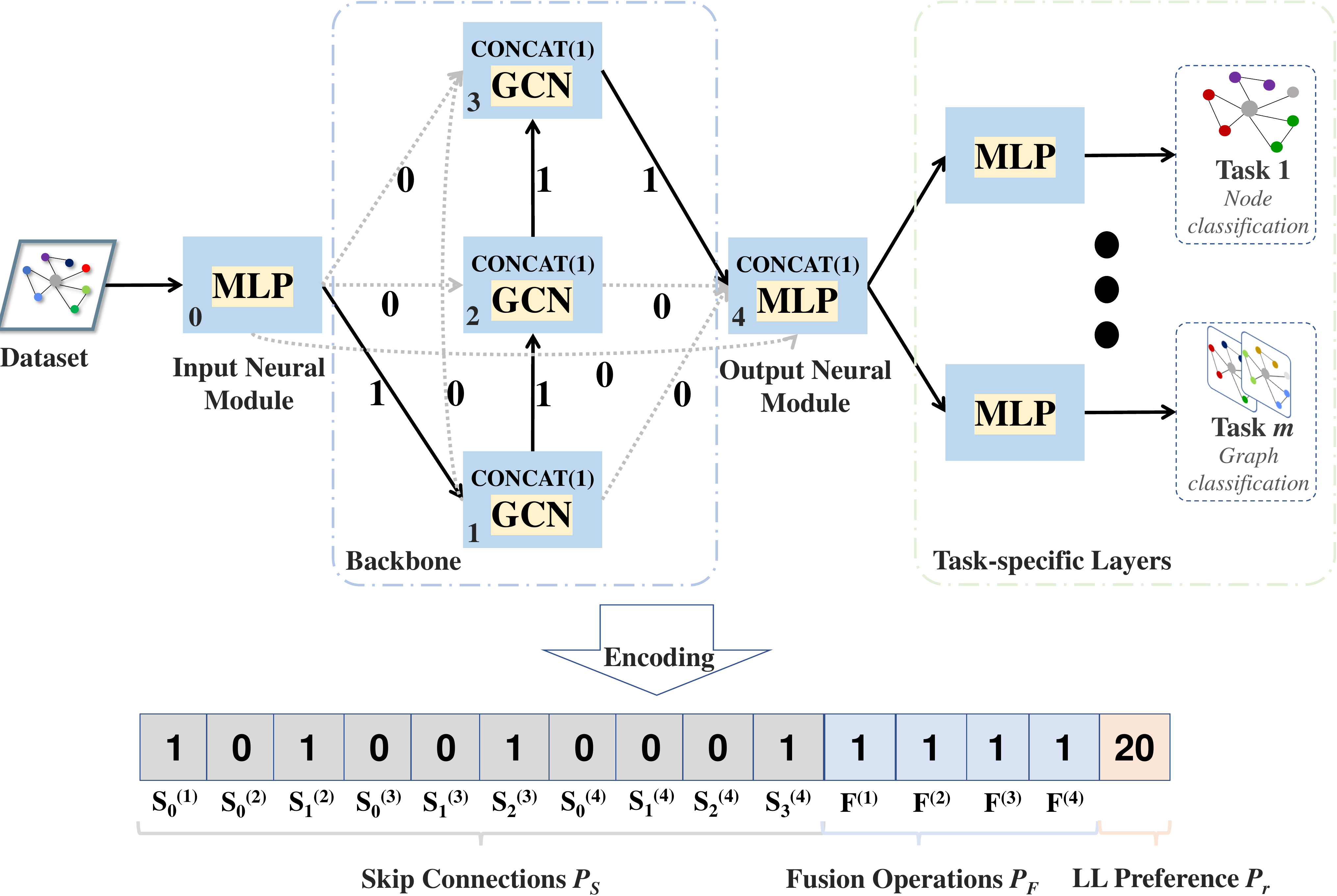}
\caption{An illustration of encoding in BLMOL for a vanill MTGCN with the three-layer backbone and CONCAT operation.} \label{encoding_GCN}
\end{figure}

According to Eq. \ref{eq7}, MTGNTS is a typical BL-MOP. So we adopt the proposed BLMOL framework (Algorithm \ref{alg2}) as the search strategy to solve the MTGNTS. In the BLMOL, a MTGNN architecture (UL variable) can be represented by an integer chromosome as shown in Fig. \ref{encoding_GCN}, which consists of skip connections $\mathrm{S}$, fusion operations $\mathrm{F}$, and LL preference vector $\bm{r}$:
\begin{equation}\label{encoding}
    P = \{ P_S, P_F, P_r\}.
\end{equation}

\textcolor{red}{We restrict the selection of $\bm{r}$ from a set of evenly distributed preference vectors $\mathcal{R}=\{\bm{r}^1,...,\bm{r}^{rm}\}$\cite{4358754}. In the chromosome, $P_r\in[1,rm]$ denotes the index of the selected preference vector in the set $\{\bm{r}^1,...,\bm{r}^{rm}\}$. $rm$ is set to 20 in this paper. For a three-layer backbone, the encoding length $D$ is $\sum_{i=1}^{4}i+4+1=15$.}

\section{Experimental Studies}
This section describes a series of experimental studies on MTGNTS to verify the efficacy of our proposed BLMOL. First, we experimentally illustrate the necessity of MTGNTS with respect to the two levels - MTL and GNAS. Then, the effectiveness of the decision-making process of LL-MOP and the preference surrogate model in BLMOL are investigated. Next, the transferability of embeddings obtained by BLMOL is introduced. For all experiments, we employ the same hyperparameter configuration (see Table \ref{Parameter-settings}) for BLMOL, where $D$ is the encoding length. Finally, these hyperparameter configurations are further analyzed.

\begin{table}[h]
\centering
\caption{Hyperparameter configurations.}\label{Parameter-settings}
\begin{tabular}{c|c|c}
\toprule
Hyperparameter & Value & Description\\
\hline
$N_s$ & 50 & Number of samples \\ 
$T$ & 500 & Number of iterations \\ 
$N_p$ & 100 & Population size \\ 
$P_c$ & 1 & Crossover probability \\ 
$P_m$ & $1/D$ & Mutation probability \\ 
\bottomrule
\end{tabular}
\end{table}

\subsection{Experiment Setting}
\paragraph{Datasets} As shown in Table \ref{datasets}, We employ three popular real-world datasets (ENZYMES, PROTEINS, and DHFR) with graph labels, node attributes, and node labels from the TUDataset library \cite{morris2020tudataset}. The ENZYMES and PROTEINS datasets are protein sets from the BRENDA database \cite{schomburg2004brenda} and the Protein Data Bank \cite{dobson2003distinguishing}, respectively. In these datasets, Nodes that represent secondary structure elements can be divided into three categories: helix, sheet, and turn. Node attributes represent physical and chemical measurements. At the graph level, the ENZYMES dataset is divided into 6 top-level classes, while the PROTEINS dataset is divided into enzyme and non-enzyme. The DHFR dataset\cite{sutherland2003spline} contains a set of compounds, which is collected from research on inhibitors of dihydrofolate reductase. The study aims to predict their toxicity. Node attributes are the 3-D coordinates of the atoms. The average number of nodes for the three datasets is 32.63, 39.06, and 42.43, while the average number of edges is 62.14, 72.82, and 44.54, respectively. These datasets satisfy multi-task conditions (node classification, graph classification, and link prediction) execution\cite{buffelli2021graph}. In each experiment, we randomly split the graph dataset into 70\%, 10\%, and 20\% for training $\mathcal{D}_{tra}$, validation $\mathcal{D}_{val}$, and testing $\mathcal{D}_{tet}$.

\begin{table}[h]
\centering
\caption{Details of datasets.}\label{datasets}
\begin{tabular}{c|c|c|c}
\toprule
Dataset &  ENZYMES & PROTEINS & DHFR \\
\hline
Number of graph & 600 & 1113 & 756 \\
Avg. nodes & 32.63 & 39.06 & 	42.43 \\
Avg. edges & 62.14 & 72.82 & 44.54 \\
Classes (graph) & 6 & 2 & 2 \\
Classes (node) & 3 & 3 & 9 \\
Node attributes & 18 & 29 & 3 \\
\bottomrule
\end{tabular}
\end{table}

\paragraph{Algorithms} EPO\cite{mahapatra2020multi}, a popular GPMOA, is embedded in BLMOL for training weights. And NSGAII\cite{deb2002fast}, a powerful MOEA, is employed to optimize UL objectives in BLMOL. The SBX crossover and PM mutation are used in NSGAII, where the distribution indexes of both SBX and PM are recommended to be set to 20 \cite{deb2002fast}.

To verify the necessity of MTGNTS and the efficacy of BLMOL, we provide four types of baselines for experiments with respect to two levels - MTL and GNAS: single-task handcrafted baselines (STHB), single-task GNAS baselines (STGB), multi-task handcrafted baselines (MTHB), and multi-task GNAS baselines (MTGB). All baselines use an encoder-decoder architecture. The encoder is the backbone to generate the embedding vector, while the decoder is the task-specific layer. In the handcrafted baseline, existing human-designed backbones are adopted. Whereas the GNAS baseline searches the topology of the backbone and the fusion operation. The decoder of the single-task baseline and the decoder of the multi-task baseline contain one task-specific layer and multiple task-specific layers, respectively. It is worth noting that all methods use the same predefined aggregation operations, number of backbones, and task-specific layers to ensure fair comparisons. In most existing GNAS works \cite{wang2021automated}, the number of backbones is set to four to obtain a good embedding representation. Consequently, we adopt a four-layer backbone, similar to most of the literature. Task-specific layers \cite{buffelli2021graph} are described as follows: 

For node classification, a single-layer neural network with a Softmax activation is employed, which maps embeddings to class predictions. For graph classification, a two-layer neural network is used to map embeddings to class predictions. The first one is a linear transformation with a ReLU activation. And then its output is averaged. The second one is a simple neural network layer with a Softmax activation. For link prediction, a two-layer neural network is used to map embeddings to the probability of a link between nodes. The first one is a linear transformation with a ReLU activation. The second one is a single-layer neural network, whose input is the concatenation of two node embeddings and whose output is the probability of the link between them.

The four types of baselines are described as follows:

\textbf{STHB} Single-task handcrafted baselines train a GNN for each task independently. Vanilla GNNs \cite{xu2018how}, ResGCN \cite{Li_2019_ICCV}, and JK-Net\cite{xu2018representation}, three state-of-the-art backbones with different topologies, are embedded in the GNN to form ST-GNN, ST-RES, and ST-JK, respectively.

\textbf{STGB} Single-task GNAS baselines search the backbone topologies for each task independently. Two popular GNAS methods are used, including random search (RS) and F$^2$ (Differentiable method)\cite{10.1145/3485447.3512185}. They are denoted as ST-RS and ST-F$^2$, respectively. For a fair comparison, the same search space as the BLMOL is employed. For the ST-RS, the number of samples is set to 100.

\textbf{MTHB} Multi-task handcrafted baselines train an MTGNN for multiple tasks with a MOO optimization strategy. Vanilla GNNs, ResGCN, and JK-Net are embedded in the MTGNN to form MT-GNN, MT-RES, and MT-JK, respectively. EPO\cite{mahapatra2020multi}, as a popular GPMOA, is adopted.

\textbf{MTGB} Multi-task GNAS baselines search the backbone topologies for multiple tasks with BLMOL. A weighted sum method is embedded in BLMOL to form BLMOL-WS, whose LL-MOP aims to minimize a weighted linear combination of LL objectives, i.e. $\mathcal{F} = \mathcal{F}_1 + \cdots + \mathcal{F}_{m}$.

\paragraph{Evaluation Metrics} Two measurement indices are employed to evaluate the performance of all methods, accuracy for NC and GC tasks and AUC for the LP task.

\paragraph{Implementation Details} More implementation details of all methods are given as follows:

All experiments are running with PyTorch (version 1.10.1), PyTorch Geometric\footnote{\url{https://github.com/pyg-team/pytorch\_geometric}} (version 2.0.3), and Geatpy (version 2.7.0) \footnote{\url{https://github.com/geatpy-dev/geatpy}} on a GPU 2080Ti (Memory: 12GB, Cuda version:11.3). The implementation of BLMOL can be accessed here \footnote{\url{https://github.com/xiaofangxd/BLMOL}}.

The batch size is set to 256 for all baselines. Models are trained with 1000 epochs for all baselines. Following existing works \cite{9458743,10.1145/3485447.3512185}, Adam with a learning rate of 0.001 is adopted. The embedding size is 256. For MTGB, in this search phase, we set the embedding size and epoch to 64 and 100 for the sake of computational resources. For STGB, the hyperparameters are the same as the original paper \cite{10.1145/3485447.3512185} for the best performance.

For MTHB, a predefined uniform preference $\bm{r}=(0.3,0.3,0.3)$ is used in EPO\footnote{\url{https://github.com/dbmptr/EPOSearch}}. We run all GNAS baselines (STGB and MTGB) fifteen times with different random seeds. For MTGB, we select the topology closest to $\bm{r}=(0.3,0.3,0.3)$ in the objective space from the final archive. Following \cite{10.1145/3485447.3512185}, the searched topologies and all handcrafted baselines (STHB and MTHB) are trained five times independently to further boost the performance. The final average performance of all methods on the test set is reported.

\subsection{Necessity of MTGNTS}
\begin{table*}
\begin{center}
\caption{Performance comparisons of all compared methods. The average accuracy rank of each method is calculated on each task.}
\label{Necessity-of-MTGNTS}
\renewcommand\arraystretch{1.25}
\LARGE
\resizebox{\textwidth}{!}{
\begin{tabular}{c|c|c|ccc|ccc|ccc|c}
			\toprule
			\multirow{2}{*}{Agg.} & \multirow{2}{*}{Category} & \multirow{2}{*}{Methods} & \multicolumn{3}{c}{ENZYMES (Mean(Std))}\vline & \multicolumn{3}{c}{PROTEINS (Mean(Std))}\vline & \multicolumn{3}{c}{DHFR (Mean(Std))}\vline & \multirow{2}{*}{Avg. Rank}\\
			\cline{4-12}
			& & & GC & NC & LP & GC & NC & LP & GC & NC & LP & \\
			\hline
			\multirow{11}{*}{GCN} & \multirow{3}{*}{STHB} &ST-GNN & 42.78(0.0241)  & 78.12(0.0031)  & 81.95(0.0004) & 73.12(0.0026) & 69.02(0.0027) & 79.16(0.0018) & 68.67(0.0581) & 95.78(0.0037) & 96.95(0.0018) & 7.78\\
			& &ST-RES & 45.56(0.0293) & 85.94(0.0024) & 86.81(0.0004) & 74.17(0.0094) & 78.04(0.0198) & 80.90(0.0069) & 71.78(0.0301) & 96.38(0.0018) & 97.48(0.0031) &  4.89\\
			& &ST-JK & 46.94(0.0315) & 86.19(0.0114) & 86.61(0.0001) & 73.12(0.0094) & 79.83(0.0090) & 82.01(0.0045) & 71.56(0.0204) & 96.94(0.0029) & 97.42(0.0016) &  4.89 \\
			\cline{2-12}
			& \multirow{2}{*}{STGB} & ST-RS & 47.63(0.0297) &  84.48(0.0124) & 86.10(0.0292) & 75.40(0.0374) & 85.12(0.0031) & 84.12(0.0023) & 72.08(0.0282) & 97.05(0.0032) & 99.03(0.0031) &  3.56 \\
			& &ST-F$^2$ & 50.24(0.0191) & \textbf{89.71(0.0039)} & 86.78(0.0082) & 73.93(0.0163) & 88.57(0.0116) & 82.53(0.0169) & 67.90(0.0319) & 97.17(0.0069) & 96.26(0.0048) & 4.11 \\
			\cline{2-12}
			& \multirow{3}{*}{MTHB} & MT-GNN & 44.17(0.0382) & 78.65(0.0098) & 69.44(0.0394) & 44.17(0.0382) & 78.65(0.0098) & 69.44(0.0394) & 72.89(0.0434) & 84.77(0.0028) & 96.05(0.0099) & 7.89 \\
			& &MT-RES & 43.89(0.0268) &  80.91(0.0078) &  77.40(0.0289) &  72.37(0.0203) &  73.41(0.0698) &  81.80(0.0408) &  71.78(0.0077) &  80.30(0.0155) &  94.54(0.0038)  & 8.11 \\
			& & MT-JK & 38.06(0.0977) &  83.07(0.0121) &  76.31(0.0124) &  73.12(0.0182) &  76.54(0.0124) &  83.46(0.0174) & 69.90(0.0454) & 85.24(0.0069) & 94.77(0.0197) &  7.56 \\
			\cline{2-12}
			& \multirow{2}{*}{MTGB} & BLMOL-WS& 52.50(0.0285) & 85.64(0.053) & 74.70(0.0305) &  75.69(0.0085) &  84.29(0.0062) &  84.01(0.0151) &  72.95(0.0219) & 88.59(0.0067) & 95.09(0.0136) & 4.67 \\
			& &\textbf{BLMOL(Ours)}& \textbf{59.13(0.0264)} & 88.94(0.0197) & \textbf{87.43(0.0055)} & \textbf{76.71(0.0391)} &  \textbf{90.79(0.0051)} &  \textbf{88.88(0.0038)} & \textbf{76.07(0.0540)} & \textbf{97.27(0.0465)} & \textbf{99.74(0.0075)} & 1.11 \\
			\midrule
			\multirow{10}{*}{GAT} & \multirow{3}{*}{STHB} &ST-GNN & 44.72(0.0459) &  83.82(0.0060) & 61.94(0.0267) & 69.82(0.0045) & 68.83(0.0204) & 81.41(0.0055) & 69.56(0.0234) & 60.38(0.0159) & 95.77(0.0059) & 7.44 \\
			& &ST-RES & 52.50(0.0363) & 87.64(0.0054) & 74.36(0.0141) & 75.08(0.0394) & 79.26(0.0023) & 80.53(0.0014) & 71.56(0.0367) & 64.67(0.0217) & 96.39(0.0040) & 5.11 \\
			& &ST-JK & 50.56(0.0488) & \textbf{87.74(0.0016)} & 80.25(0.0089) & 76.58(0.0295) & 80.13(0.0029) & 81.06(0.0052) & 68.44(0.0336) & 51.51(0.0137) & 96.81(0.0032)& 5.00  \\
			\cline{2-12}
			& \multirow{2}{*}{STGB} & ST-RS & 53.09(0.0560) & 84.36(0.0015) & 75.87(0.0272) & 74.24(0.0236) & 81.32(0.0143) & 82.19(0.0070) & 70.89(0.0577) & 69.28(0.0255) & 97.35(0.0059) & 4.33  \\
			& &ST-F$^2$ & 52.36(0.0260) & 86.65(0.0221) & 75.36(0.0360) & 75.61(0.0232) & 84.45(0.0130) & 81.62(0.0113) & 71.55(0.0135) &  81.21(0.0239) & 97.80(0.0074) &  3.44 \\
			\cline{2-12}
			& \multirow{3}{*}{MTHB} & MT-GNN & 56.67(0.0500) &  66.69(0.0137) & 60.62(0.0128) & 75.53(0.0182) & 57.92(0.0178) & 77.52(0.0180) &  67.78(0.0336) &  54.77(0.0069) &  94.61(0.0061) & 8.11  \\
			& &MT-RES & 54.17(0.0083) &  79.63(0.0084)& 71.94(0.0113) & 70.57(0.0496) & 67.83(0.0522) & 79.12(0.0121) & 66.00(0.0267) & 59.05(0.0271) & 96.70(0.0034) & 7.56 \\
			& & MT-JK &  51.67(0.0520) & 85.23(0.0103) & 77.37(0.0074) & 76.13(0.0045) & 69.79(0.0264) & 81.18(0.0043) & 68.89(0.0077) & 56.70(0.0516) & 95.79(0.0063) & 5.78 \\
			\cline{2-12}
			& \multirow{2}{*}{MTGB} & BLMOL-WS & 55.75(0.0235) & 85.35(0.0157) & 74.95(0.0034) &  73.73(0.0488) &  59.74(0.0287) &  80.34(0.0212) & 65.11(0.0222) & 74.15(0.0158) & 93.89(0.0170) & 6.78 \\
			& &\textbf{BLMOL(Ours)} & \textbf{58.84(0.0214)} &  85.31(0.0064) & \textbf{82.91(0.0209)} & \textbf{77.31(0.0172)} &  \textbf{86.22(0.0288)} & \textbf{85.33(0.0383)} & \textbf{75.51(0.0041)} &  \textbf{86.61(0.0015)} & \textbf{97.83(0.0054)} & 1.44 \\
			\bottomrule
\end{tabular}
}
\end{center}
\end{table*}

Table \ref{Necessity-of-MTGNTS} lists the experimental results in terms of accuracy (\%) for NC and GC and AUC (\%) for LP obtained by all comparative methods based on predefined aggregation operations. Two popular aggregation operations are considered, including GCN and GAT. We report the average value and the standard deviation. The best performances in each group are indicated in bold. The average accuracy rank of each method is calculated on each task. First, handcrafted baselines with different topologies have different performances. Compared to simple stacking (STGNN), GNNs with residuals (ST-RES and ST-JK) exhibit better performance in terms of the average accuracy rank. This illustrates the importance of studying topology. Second, the overall performance of all baselines (STGB and MTGB) that automatically search for topology and fusion strategy is better than that of handcrafted baselines (STHB and MTHB). The performance gain indicates that the searched topology has a stronger representation ability than the simple stacked handcrafted graph network, which validates the necessity of studying the GNAS method (bi-level optimization). Third, MTHB does not achieve a huge advantage over STHB due to inter-task conflict. A handcrafted baseline with limited representation ability cannot efficiently handle conflicts between tasks, leading to performance degradation. Fourth, BLMOL has better overall performance than STGB, which benefits from real-time trade-offs and knowledge sharing between objectives. This indicates the usefulness of the BLMOL in designing the topology and the necessity of studying MOO for MTL. Finally, the overall performance of BLMOL outperforms BLMOL-WS, which shows that our proposed decision method for LL-MOP is more effective than the weight sum method.

The total search cost consumed by the GNAS baseline can be divided into two stages: (1) Pre-search: the search cost incurred prior to the search, such as the construction of a preference surrogate model in BLMOL. (2) During-search: the search cost of executing the search algorithm. Table \ref{search-time} lists the averaged search time of GNAS baselines on different datasets over fifteen independent runs. "–" indicates not applicable. For STGB, search times on all tasks are accumulated. The average accuracy rank of each method is calculated on each dataset. 

\begin{table}[t]
\begin{center}
\renewcommand\arraystretch{1.25}
\LARGE
\caption{Comparing the search cost (GPU second) of BLMOL to other GNAS baselines. The average accuracy rank of each method is calculated on each dataset.}\label{search-time}
\resizebox{\linewidth}{!}{ 
\begin{tabular}{c|c|c|c|c|ccc|c}
\toprule
\multirow{2}{*}{Agg.} & \multirow{2}{*}{Dataset} & \multirow{2}{*}{GNAS Baselines} & \multirow{2}{*}{Avg. Rank} & \multirow{2}{*}{Pre-Search} &  \multicolumn{3}{c}{During-Search}\vline & \multirow{2}{*}{Total} \\
\cline{6-8}
 &  &  &  & & GC & NC & LP & \\
\hline
\multirow{12}{*}{GCN} & \multirow{4}{*}{ENZYMES} & ST-RS & 5.00 & - & 17.1k & 23.1k & 22.2k & 62.4k \\
& & ST-F$^2$ (gradient) & 2.33
 & - & 1.2k & 1.5k & 3.2k & \textcolor{red}{\textbf{5.9k}} \\
& & BLMOL-WS& 5.33 & 24.1k & \multicolumn{3}{c}{7.0}\vline & 24.1k\\
& & \textbf{BLMOL(Ours)}& \textcolor{red}{\textbf{1.33}} & 20.4k & \multicolumn{3}{c}{5.9}\vline & 20.4k\\
\cline{2-9}
& \multirow{4}{*}{PROTEINS} & ST-RS & 2.67 & - & 13.6k & 15.4k & 47.7k & 76.7k \\
& & ST-F$^2$ (gradient) & 4.00 & - & 2.7k & 2.5k & 6.8k & \textcolor{red}{\textbf{12.0k}} \\
& & BLMOL-WS & 3.00 & 35.1k & \multicolumn{3}{c}{9.4}\vline & 35.1k\\
& & \textbf{BLMOL(Ours)}& \textcolor{red}{\textbf{1.00}} & 36.8k & \multicolumn{3}{c}{9.9}\vline & 36.8k \\
\cline{2-9}
& \multirow{4}{*}{DHFR} & ST-RS & 3.00 & - & 18.8k & 12.4k & 26.2k & 57.4k \\
& & ST-F$^2$ (gradient) & 6.00 & - & 0.8k & 1.2k & 1.8k & \textcolor{red}{\textbf{3.8k}} \\
& & BLMOL-WS & 5.67 & 30.2k & \multicolumn{3}{c}{10.6}\vline & 30.2k \\
& & \textbf{BLMOL(Ours)} & \textcolor{red}{\textbf{1.00}} & 30.9k & \multicolumn{3}{c}{9.0}\vline & 30.9k \\
\midrule
\multirow{12}{*}{GAT} & \multirow{4}{*}{ENZYMES} & ST-RS & 5.33 & - & 19.7k & 24.0k & 17.8k & 61.5k \\
& & ST-F$^2$ (gradient) & 5.00 &- & 1.3k & 1.7k & 2.6k & \textcolor{red}{\textbf{5.6k}}\\
& & BLMOL-WS& 4.33 & 24.3k & \multicolumn{3}{c}{6.6}\vline & 24.3k \\
& & \textbf{BLMOL(Ours)} & \textcolor{red}{\textbf{2.33 }}& 24.4k & \multicolumn{3}{c}{7.6}\vline & 24.4k \\
\cline{2-9}
& \multirow{4}{*}{PROTEINS} & ST-RS &4.00& - & 13.5k & 11.9k& 34.8k & 60.2k \\
& & ST-F$^2$ (gradient) & 3.00 &- & 2.6k & 2.1k & 7.2k & \textcolor{red}{\textbf{11.9k}} \\
& & BLMOL-WS & 8.33 & 37.4k & \multicolumn{3}{c}{9.8}\vline & 37.4k \\
& & \textbf{BLMOL(Ours)} & \textcolor{red}{\textbf{1.00}} & 38.0k & \multicolumn{3}{c}{10.2}\vline & 38.0k \\
\cline{2-9}
& \multirow{4}{*}{DHFR} & ST-RS & 3.67 &- & 11.8k & 12.0k & 30.0k & 53.8k \\
& & ST-F$^2$ (gradient) & 2.33 & - & 0.7k & 1.0k & 1.6k & \textcolor{red}{\textbf{3.3k}} \\
& & BLMOL-WS & 7.67 & 29.7k & \multicolumn{3}{c}{10.1}\vline & 29.7k \\
& & \textbf{BLMOL(Ours)} & \textcolor{red}{\textbf{1.00}} & 36.2k & \multicolumn{3}{c}{9.6}\vline & 36.2k \\
\bottomrule
\end{tabular}
}
\end{center}
\end{table}

\begin{figure*}[ht]
	\centering
	\subfloat[ENZYMES-GCN]{
		\includegraphics[width=0.3\linewidth]{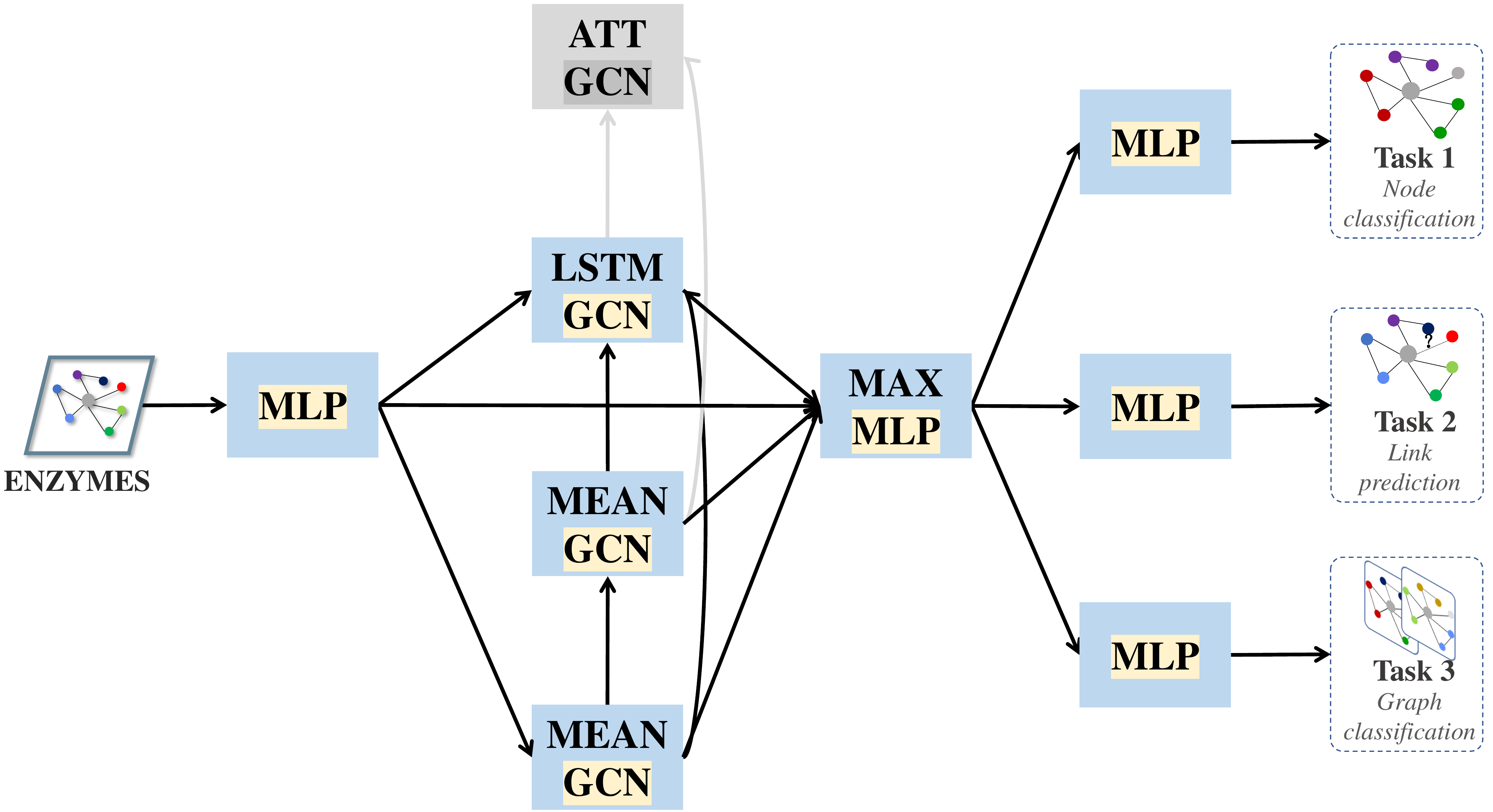}}
	\hfill
	\subfloat[ENZYMES-GAT]{
		\includegraphics[width=0.3\linewidth]{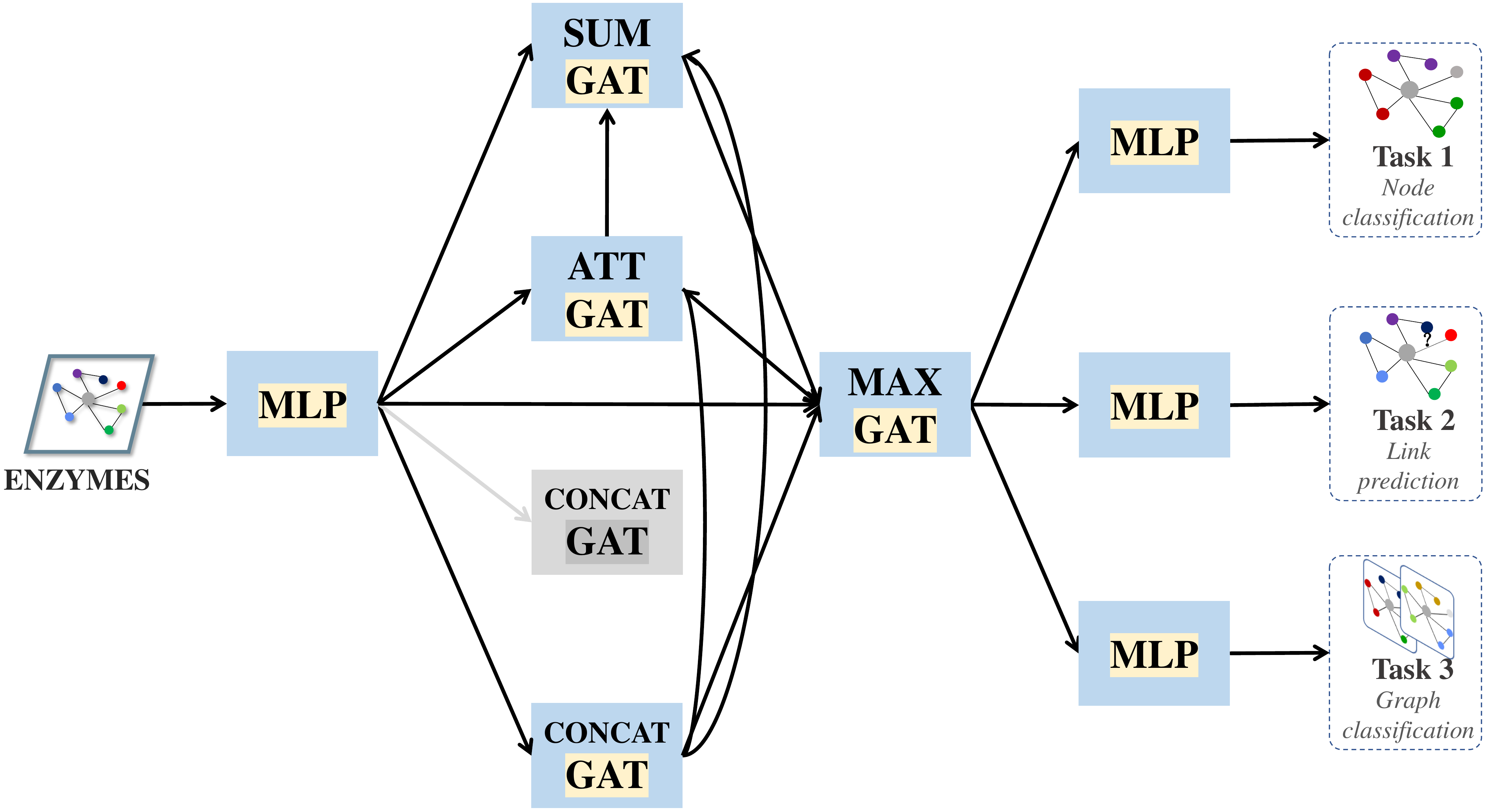}}
	\hfill
	\subfloat[PROTEINS-GCN]{
		\includegraphics[width=0.3\linewidth]{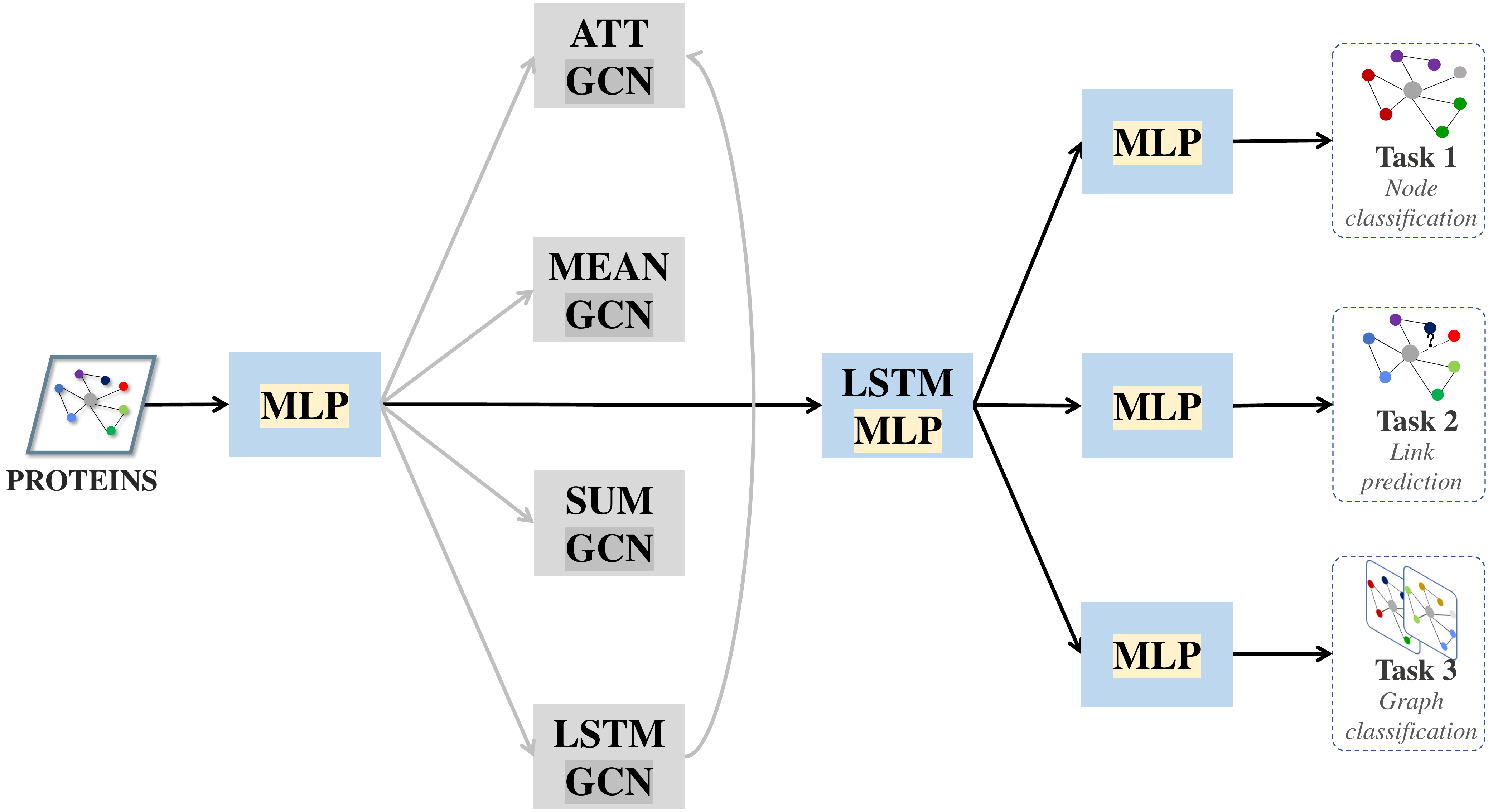}}
	\hfill
	\subfloat[PROTEINS-GAT]{
		\includegraphics[width=0.3\linewidth]{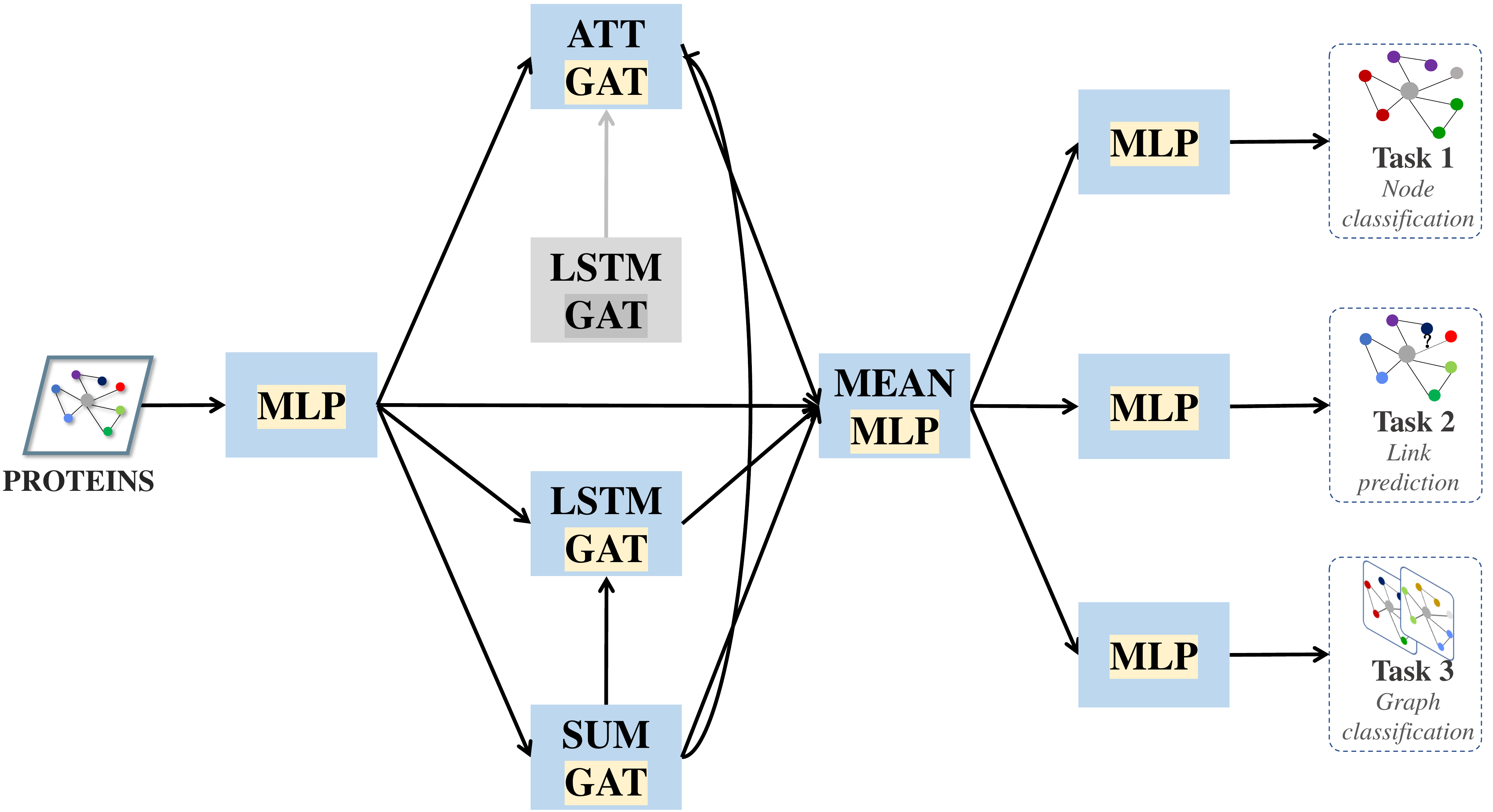}}
	\hfill
	\subfloat[DHFR-GCN]{
		\includegraphics[width=0.3\linewidth]{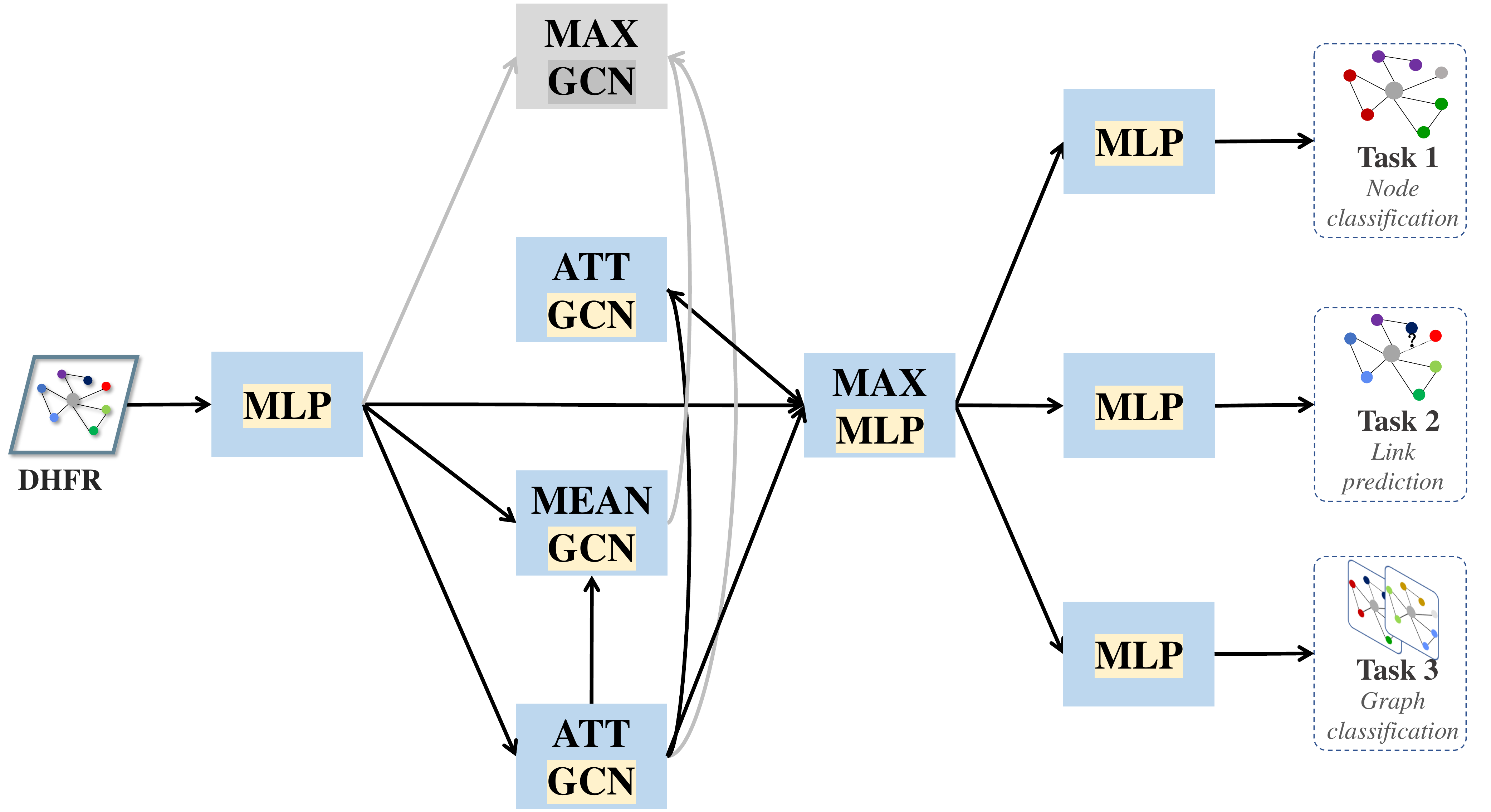}}
	\hfill
	\subfloat[DHFR-GAT]{
		\includegraphics[width=0.3\linewidth]{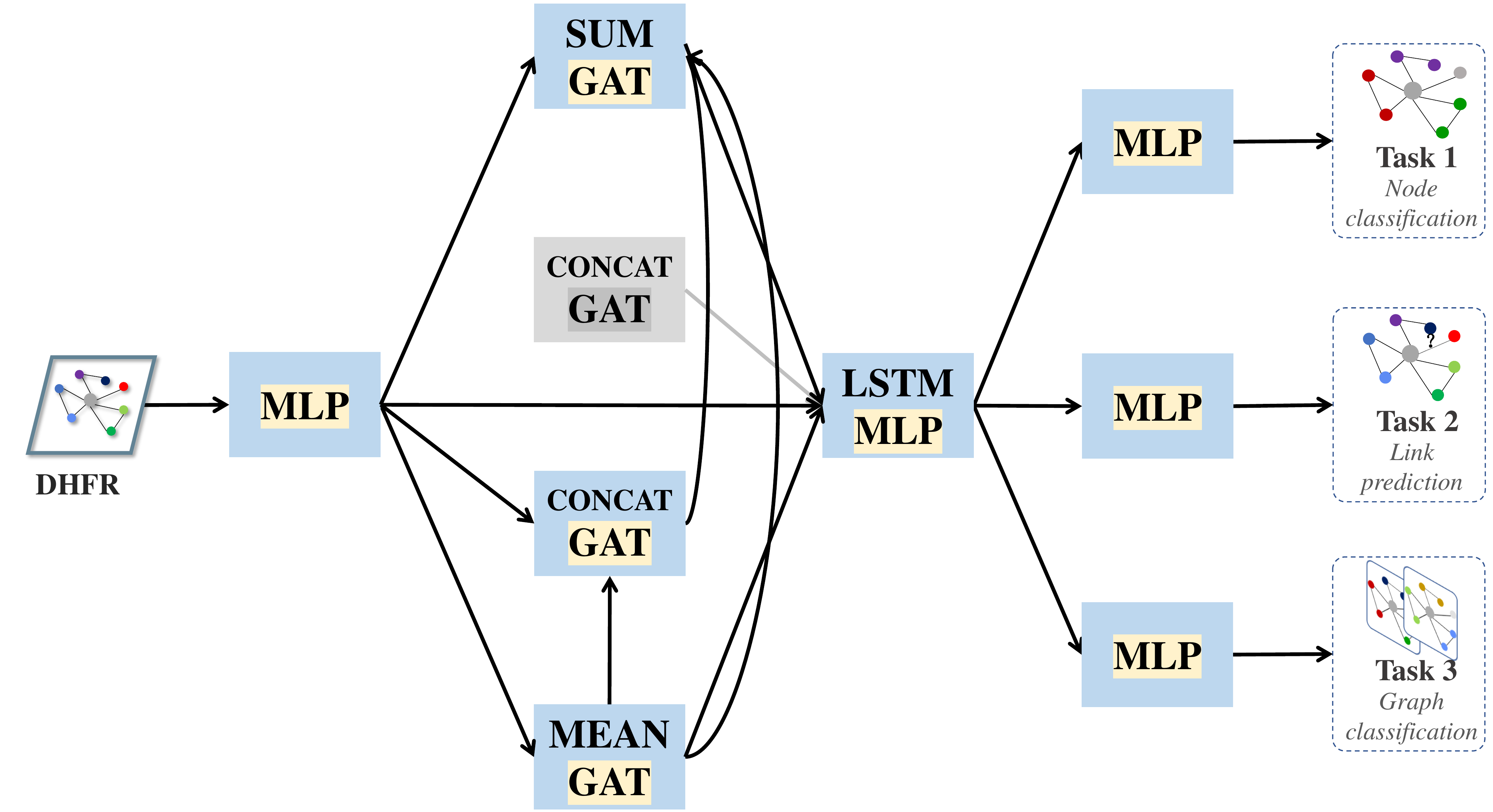}}
	\caption{Visualization of architectures searched by BLMOL on all datasets, where grey blocks indicate that they are not employed in training.}
	\label{fig_arch}
\end{figure*}

Compared with the ST-RS, BLMOL has better overall performance in less time by searching for the multi-task topology and the fusion operation, which shows the efficiency of our designed GNAS method. The differentiable method (ST-F$^2$) has a lower search cost than BLMOL. But BLMOL has higher performance in terms of the avg. rank. Again, we emphasize that BLMOL is more flexible. It does not require differentiable approximations or assumptions. In Fig. \ref{fig_arch}, we visualize architectures searched by BLMOL on different cases. For different aggregation operations and datasets, the searched topologies are significantly different, which further illustrates the importance of the GNAS method. Furthermore, the initial features generated by the input block are used in the output block in almost all topologies. These initial features may contain more information about the nodes themselves. This conclusion is consistent with the studies in \cite{chen2020simple, NEURIPS2020_58ae23d8, 10.1145/3485447.3512185}.

\subsection{Effectiveness of Decision-making Process of LL-MOP }

\begin{figure*}[ht]
	\centering
	\subfloat[ENZYMES]{
		\includegraphics[width=0.3\linewidth]{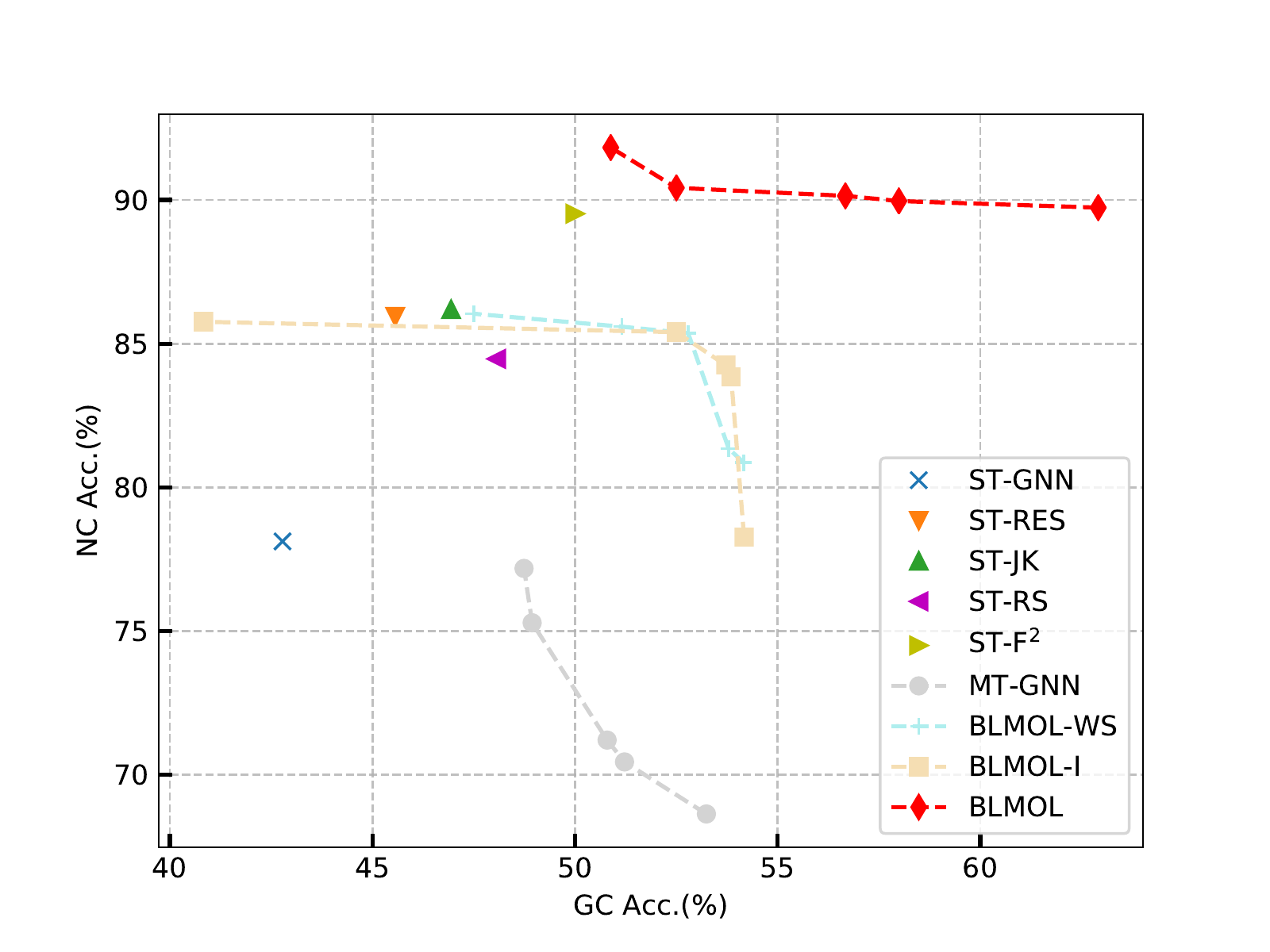}}
	\hfill
	\subfloat[PROTEINS]{
		\includegraphics[width=0.3\linewidth]{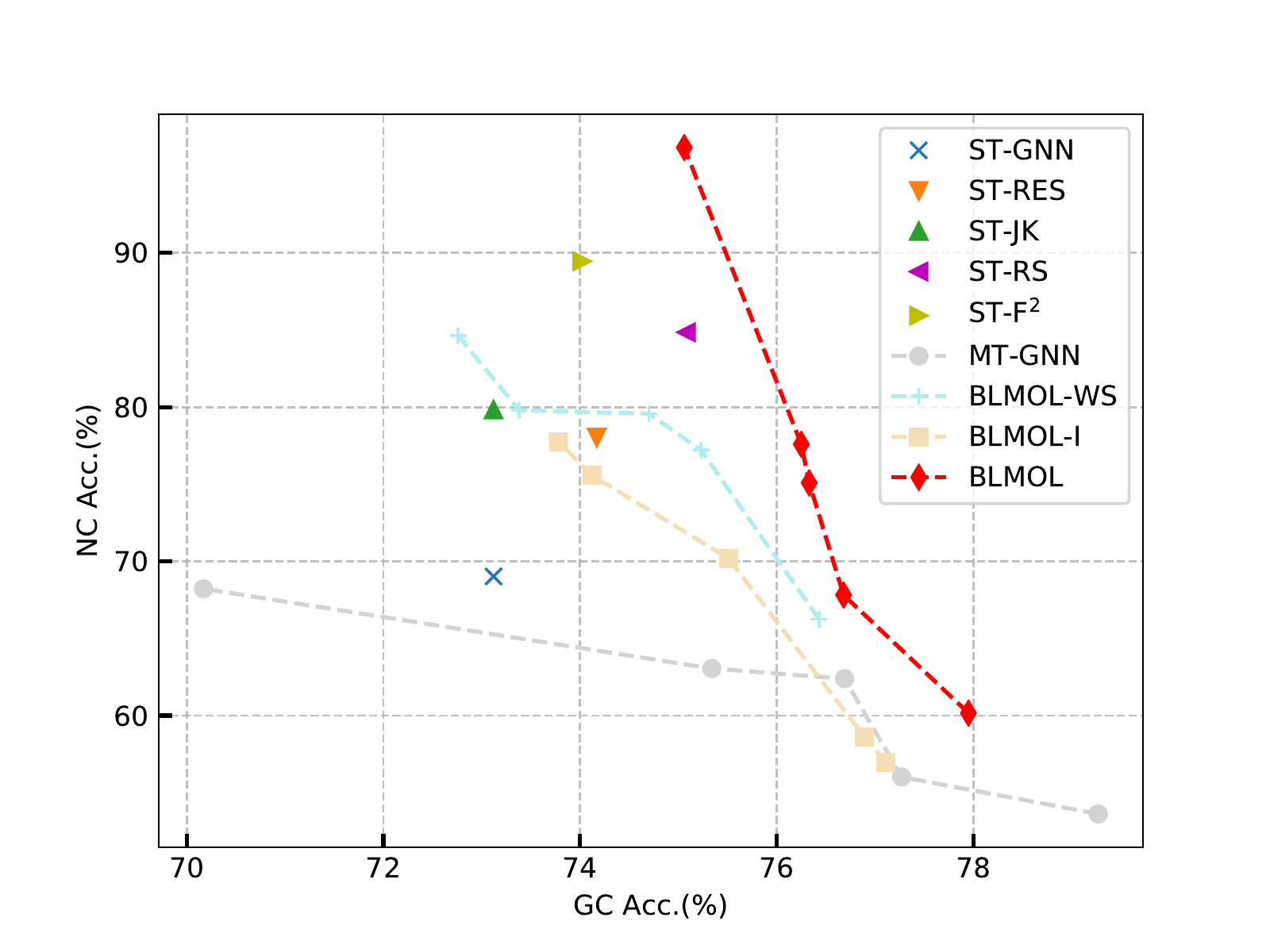}}
	\hfill
	\subfloat[DHFR]{
		\includegraphics[width=0.3\linewidth]{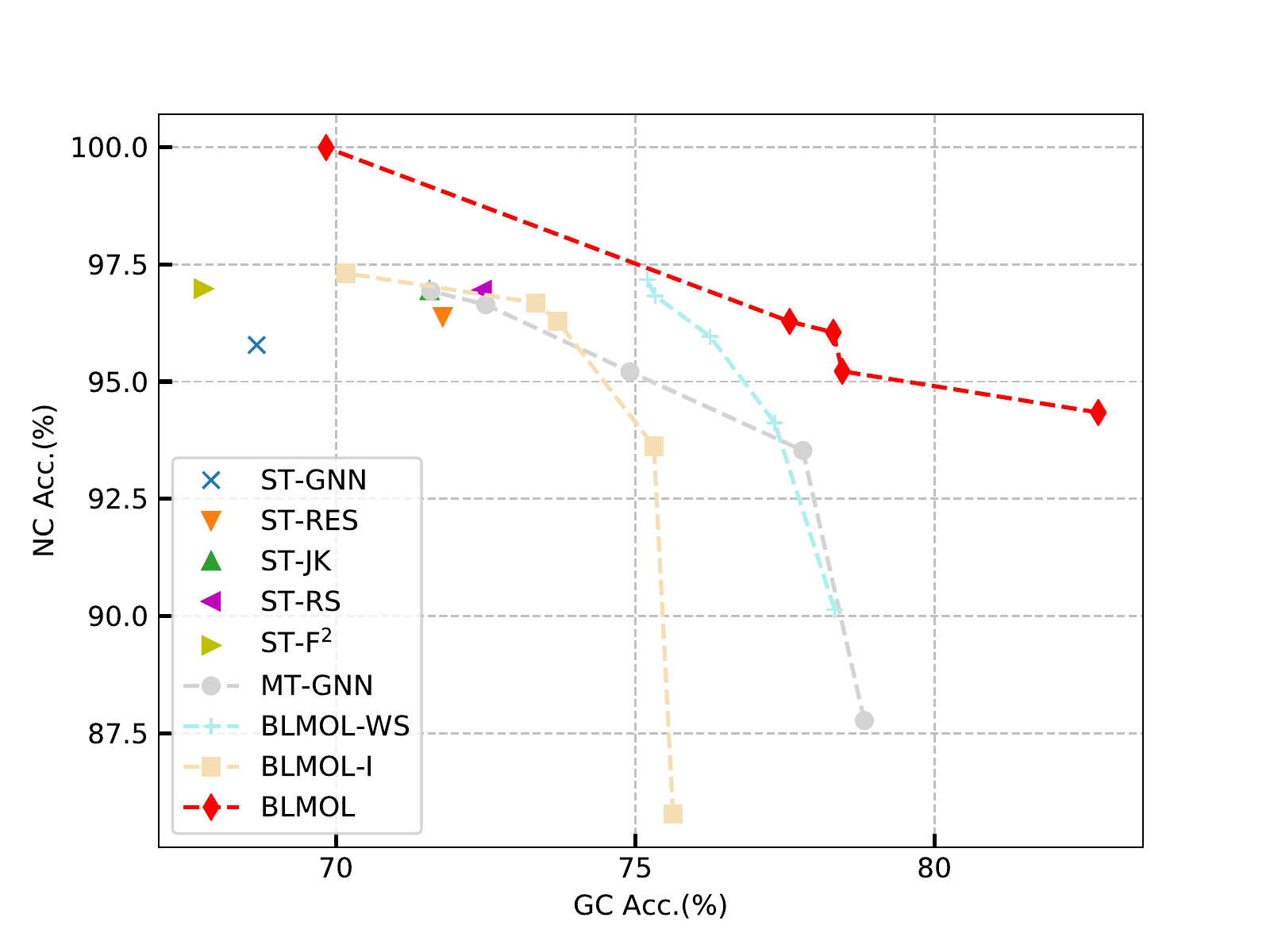}}
	\caption{Comparison of task accuracy trade-off curves of BLMOL and compared baselines. The aggregation operation is GCN.}
	\label{fig_gc_nc}
\end{figure*}

In this section, we perform a series of experimental studies on two graph classification tasks: GC and NC. A new compared baseline, called BL-MOL-I, is constructed to verify the effectiveness of the decision-making process of LL-MOP in BL-MOL, where the LL preference vector is randomly sampled from a set of even spread preference vectors $\mathcal{R}$. Fig. \ref{fig_gc_nc} shows the comparison of task accuracy trade-off curves of BLMOL and baselines. The aggregation operation is GCN. For MTHBs, five predefined preference vectors $\bm{r}$ are employed in EPO, including (0.9, 0.1), (0.75, 0.25), (0.5, 0.5), (0.25, 0.75), and (0.1, 0.9). For BLMOL and multi-task GNAS baselines, we select five topologies closest to these predefined preference vectors in the objective space from the final archive.

Experimental results indicate that BL-MOL completely dominates all existing designs, including handcrafted and GNAS methods, in terms of GC and NC. Compared to handcrafted models, automated design methods are significantly more efficient. BLMOL outperforms ST-RS and ST-F$^2$ (a state-of-the-art differentiable method), demonstrating that BLMOL can search for better-performing GNN topologies \textbf{red}{with superior performance}. For ENZYMES datasets, we observe that every GNN obtained by BLMOL completely dominates the single-task baseline, suggesting a synergy between the NC task for secondary structure elements and the protein classification task. That is, the nature of secondary structure elements may determine the type of protein. In all datasets, through knowledge sharing, multi-task baselines can achieve excellent performance on a specific task that cannot be achieved by single-task ones, suggesting that there may be a high correlation between the graph embedding information required by NC and GC in graph representation learning. Furthermore, BLMOL outperforms BLMOL-WS and BLMOL-I. BLMOL-WS simply weights multiple LL objectives while ignoring potential conflicts, resulting in poor performance. Essentially, BLMOL-I randomly selects an exact LL Pareto weight for each UL variable. However, BLMOL searches for the most suitable exact LL Pareto weight by maximizing the UL objective. Therefore, the effectiveness of our proposal decision-making process of LL-MOP is verified.

\subsection{Effectiveness of Preference Surrogate Model}
In this section, we evaluate the effectiveness of different preference surrogate models. We uniformly sample 100 samples (topology, fusion operations, and preference vectors) from the search space and train these architectures with EPO on the ENZYMES dataset for 100 epochs. We randomly select 10\% of the data for testing, and the remaining 90\% of the data is used to train the preference surrogate model. All models are implemented by scikit-learn [22] with the default parameter settings, which tend to yield satisfactory performance. Two popular metrics, KTau and MSE, are used to evaluate the prediction results.
Table \ref{PRM} shows the predicted accuracy of surrogate models on all tasks. From experimental results, none of the surrogate models outperform the others in both KTau and MSE. Therefore, in BLMOL, a surrogate model with the best performance is selected for each UL objective as the final surrogate model.

\begin{figure}
\centering
\includegraphics[width=0.4\textwidth]{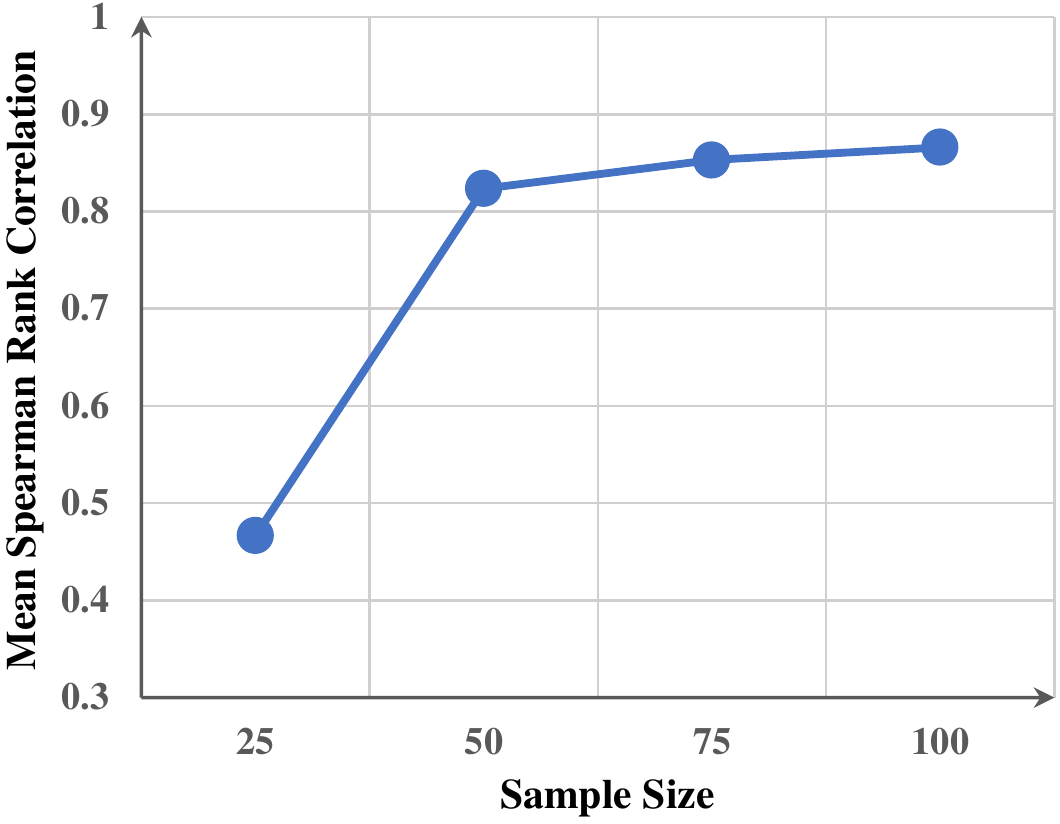}
\caption{Preference surrogate model performance as a function of the sample size on the ENZYMES dataset. For each model, we show the mean Spearman rank correlation on three tasks (GC, NC, and LP).}
\label{SRC}
\end{figure}

In addition, Fig \ref{SRC} illustrates the mean (over three tasks) Spearman rank correlation between the predicted and the true accuracy of the preference surrogate model with various configurations of the sample size $N_s$. $N_s$ is varied from 25 to 100 with an interval of 25. It can be observed that the mean Spearman rank correlation increases with the increase of the sample size. It is recommended to set the sample size to 50 for all our experiments to obtain a better compromise between minimizing the number of samples (reducing the computational cost of the Pre-search) and maximizing the Spearman rank correlation (improving the accuracy of the preference surrogate model).

\begin{table}[t]
\renewcommand\arraystretch{1.25}
\LARGE
\caption{Predicted accuracy of six surrogate models on the ENZYMES dataset.}
    \centering
    \label{PRM}
    \resizebox{\linewidth}{!}{ 
    \begin{tabular}{c|cc|cc|cc}
    \toprule
        \multirow{2}{*}{Model} & \multicolumn{2}{c}{GC}\vline & \multicolumn{2}{c}{NC}\vline & \multicolumn{2}{c}{LP}   \\
        \cline{2-7}
         & KTau & MSE & KTau & MSE & KTau & MSE  \\
         \hline
        SVR & 0.1999 & 0.0039 & \textbf{0.7999} & 0.0087 & 0.3999 & 0.0028  \\
        RF & 0.3999 & \textbf{0.0023} & 0.3999 & 0.0078 & 0.6000 & 0.0036 \\
        Adaboost & \textbf{0.6000} & 0.0072 & 0.6000 & \textbf{0.0043} & 0.3999 & 0.0033  \\
        Bagging & 0.3999 & 0.0039 & \textbf{0.7999} & 0.0076 & \textbf{0.9999} & \textbf{0.0011}  \\
        GP & -0.3999 & 0.0634 & 0.6000 & 0.6583 & 0.7999 & 0.6952  \\
        MLP & 0.7999 & 0.0767 & -0.3999 & 0.0487 & -0.7999 & 0.0696  \\
    \bottomrule
    \end{tabular}
    }
\end{table}

\subsection{Transferability}
\begin{figure*}[ht]
	\centering
	\subfloat[ENZYMES]{
		\includegraphics[width=0.3\linewidth]{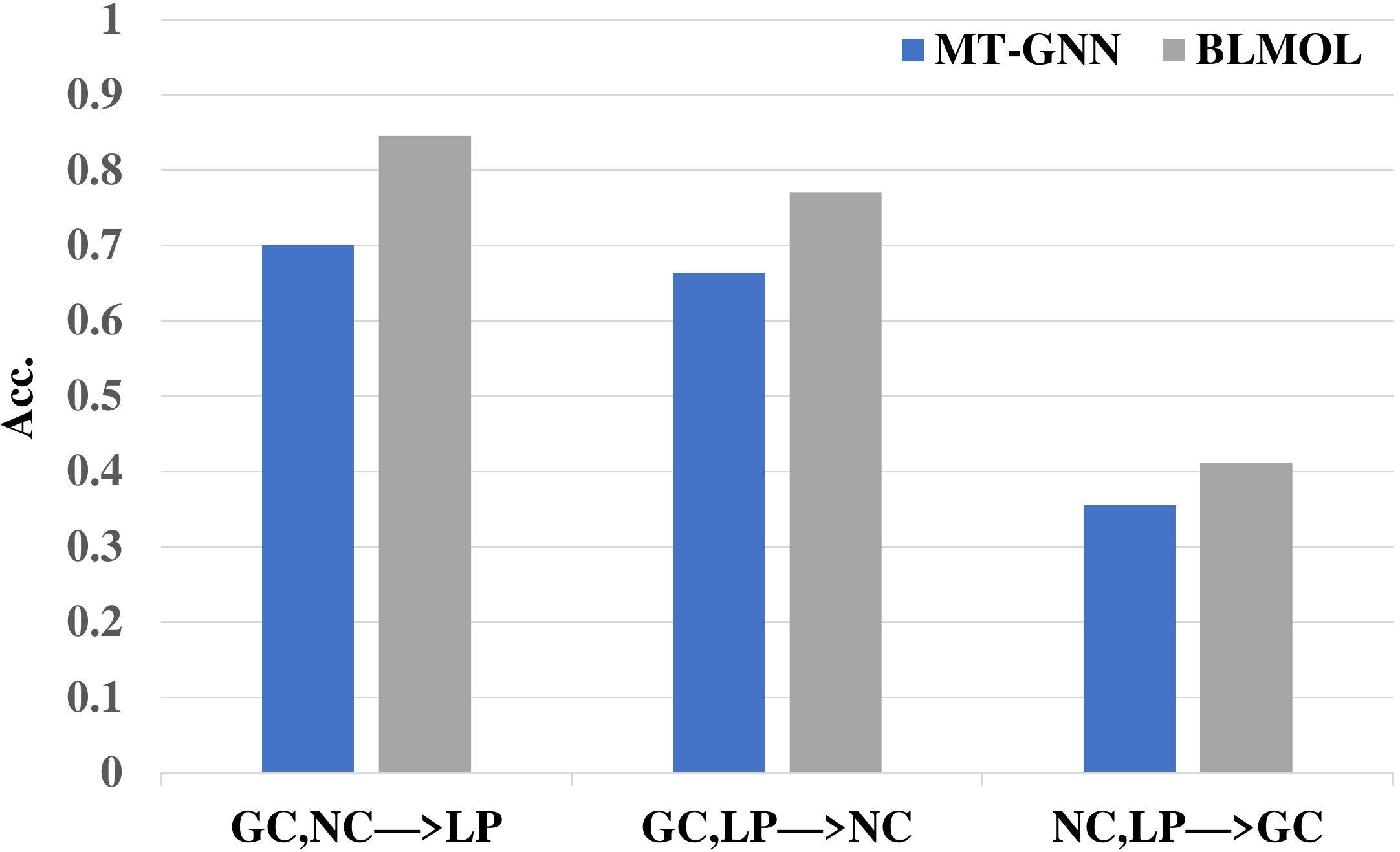}}
	\hfill
	\subfloat[PROTEINS]{
		\includegraphics[width=0.3\linewidth]{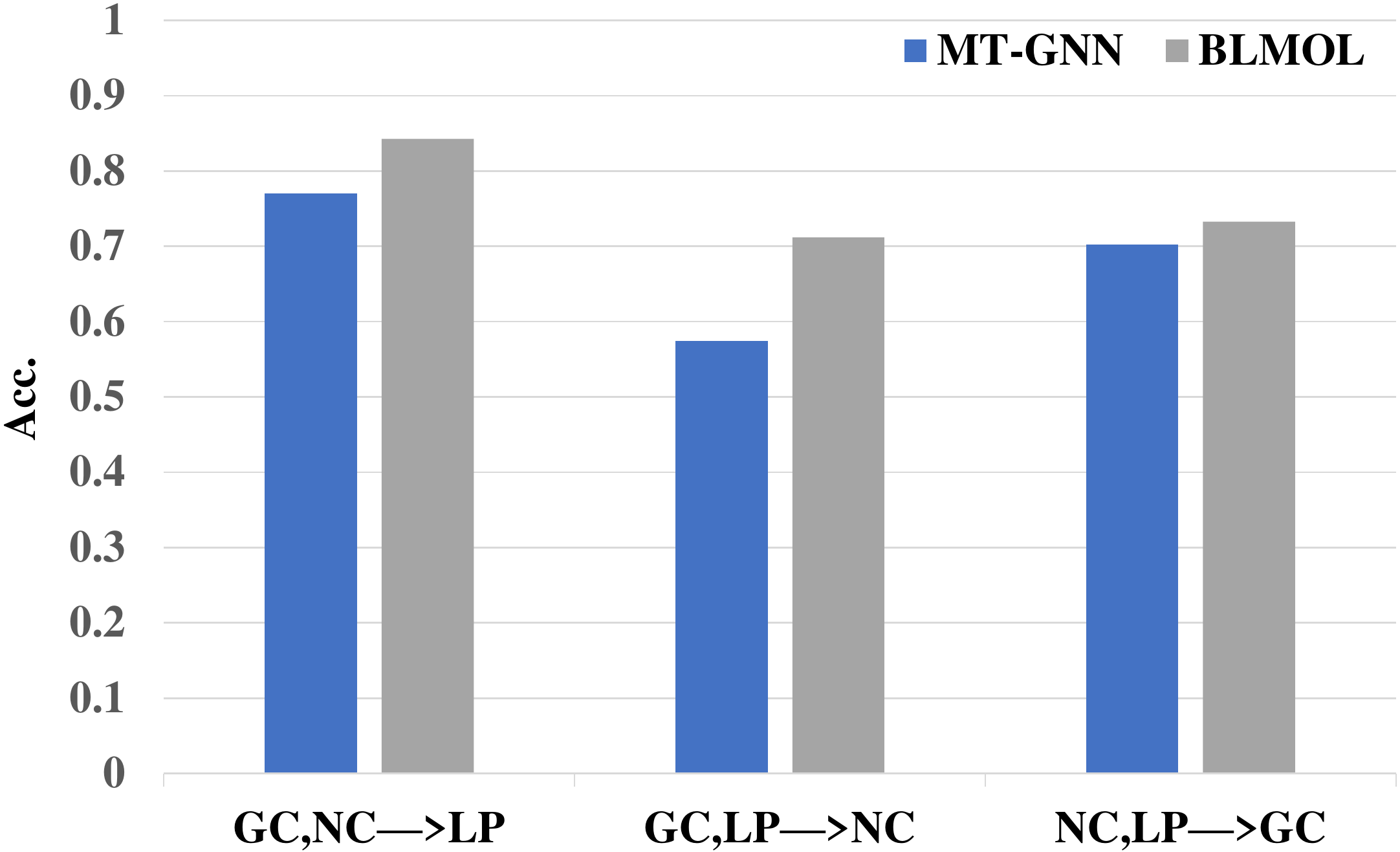}}
	\hfill
	\subfloat[DHFR]{
		\includegraphics[width=0.3\linewidth]{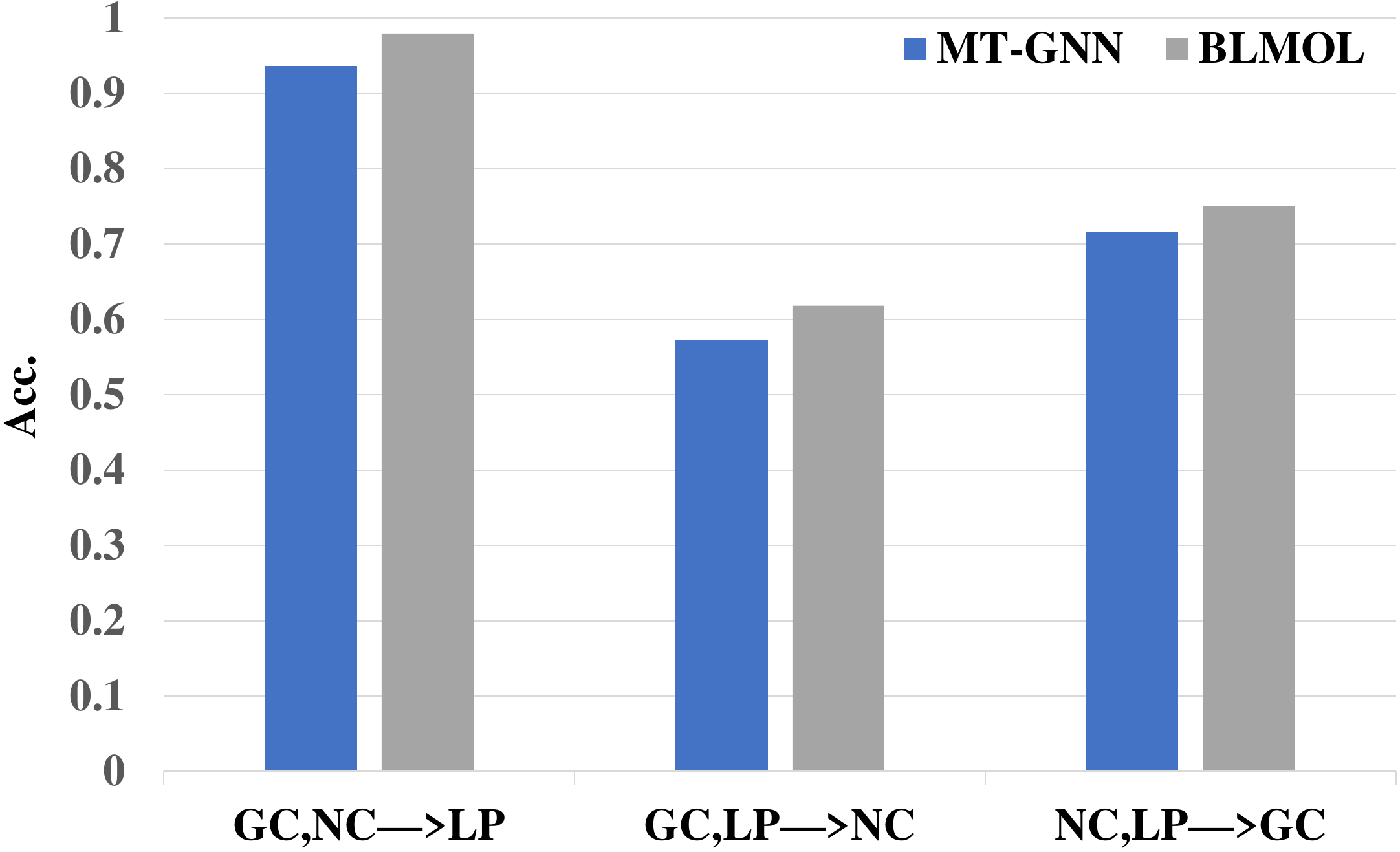}}
	\caption{Comparison of transferability of BLMOL and MT-GNN baselines. The preference vector $\bm{r}$ and aggregation operation are (0.5, 0.5) and GCN, respectively.}
	\label{transfer}
\end{figure*}

To demonstrate the transferability of our proposed BLMOL, we train a task-specific layer for unseen tasks using node embeddings obtained from MT-GNN and BLMOL on any two tasks. Here $a, b \rightarrow{c}$ means that $a$ and $b$ are the training tasks, and $c$ is the unseen task. Take the following experiment settings as an example. The preference vector $\bm{r}$ and aggregation operation are set to (0.5, 0.5) and GCN, respectively. Embedding vectors produced by the multi-task models trained on $a$ and $b$ are used as input for task-specific layers of the unseen task $c$. We employ SGD to train the task-specific layer for 100 epochs. Fig. \ref{transfer} shows the accuracy of the unseen task $c$ on all datasets. BLMOL achieves the highest accuracy on all unseen tasks than MT-GNN, indicating that our method can obtain node embedding information with higher generalization than handcrafted methods by knowledge transfer across tasks.

\subsection{Parameter Analysis}

In this section, we provide additional experiments to analyze some key parameters, including the number of backbones, the crossover probability $P_c$, and the mutation probability $P_m$.

\begin{figure*}[ht]
	\centering
	\subfloat[Task$_1$: GC]{
		\includegraphics[width=0.3\linewidth]{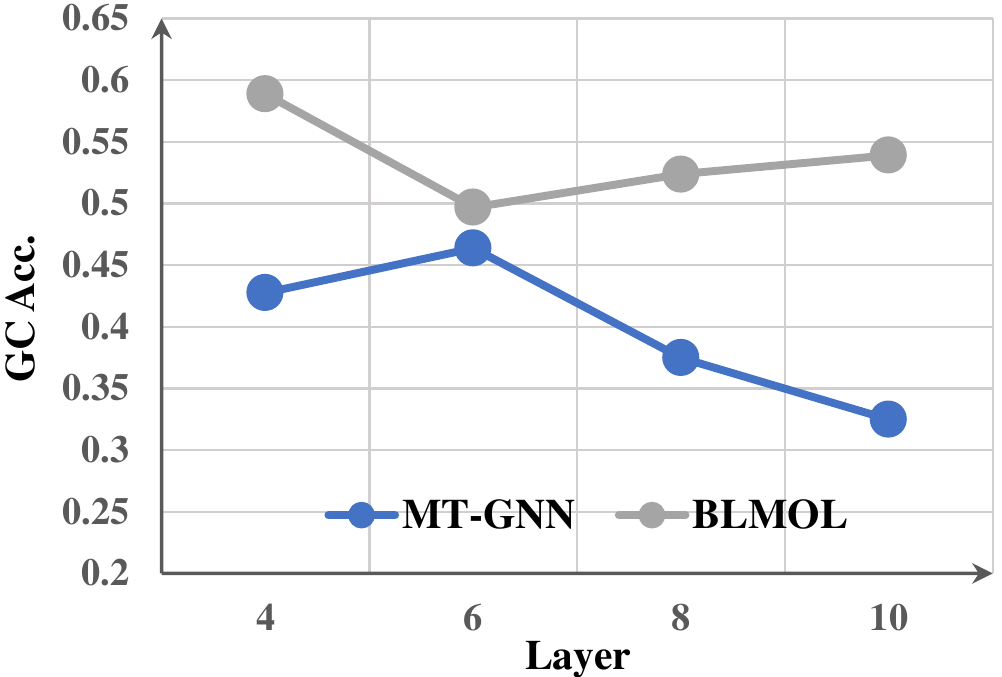}}
	\hfill
	\subfloat[Task$_2$: NC]{
		\includegraphics[width=0.3\linewidth]{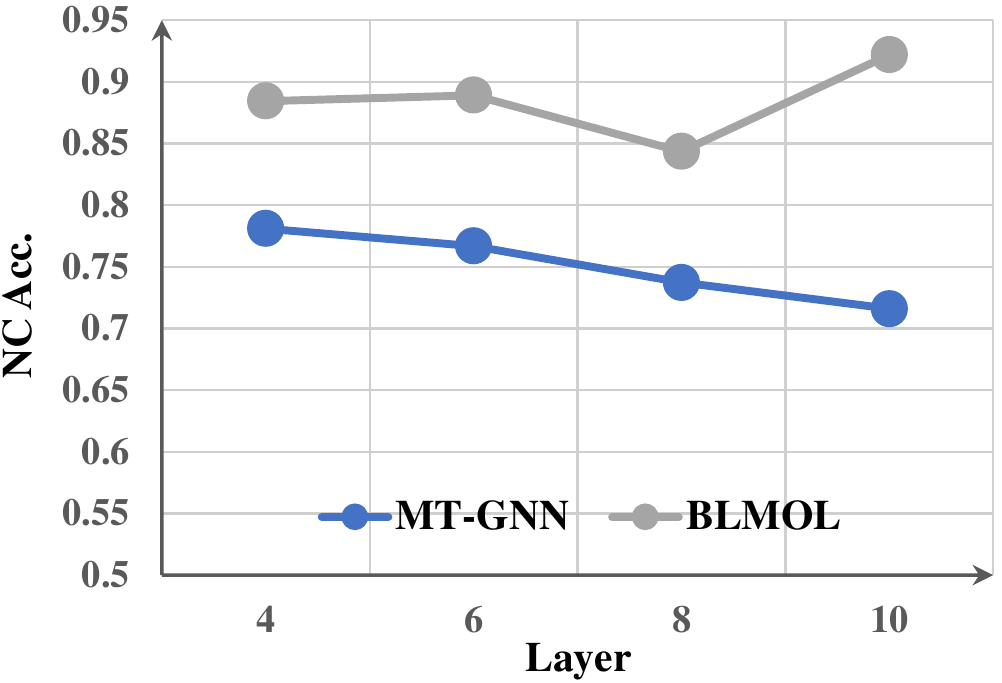}}
	\hfill
	\subfloat[Task$_3$: LP]{
		\includegraphics[width=0.3\linewidth]{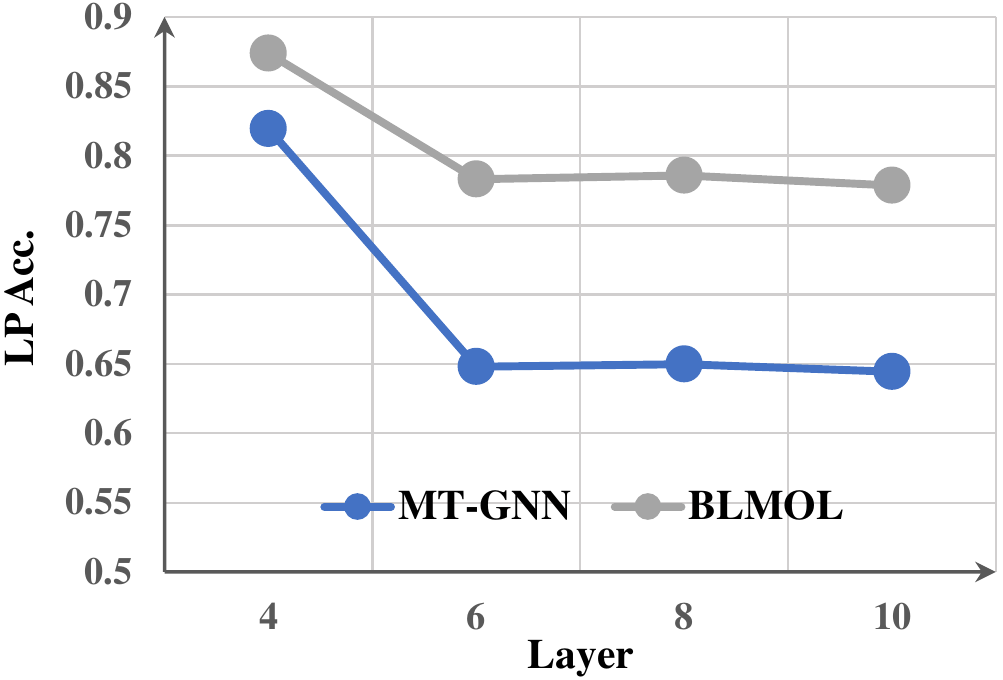}}
	\caption{Task accuracy on the ENZYMES dataset obtained by the BLMOL and MT-GNN with various configurations of parameters $N_b$. The aggregation operation is GCN.}
	\label{Layer}
\end{figure*}

First, the impact of the number of backbones $N_b$ on the performance of BLMOL is explored. Existing handcrafted models obtain higher-level features through simple stacking aggregation operations. However, as the number of backbones increases, existing GNN models suffer from over-smoothing problems, i.e. the node representation becomes indistinguishable and can easily get a performance drop \cite{10.5555/3504035.3504468}. Therefore, it is challenging to overcome the over-smoothing problem for deeper GNNs. Fig. \ref{Layer} shows the task accuracy on the ENZYMES dataset obtained by the BLMOL and MT-GNN with various configurations of the parameter $N_b$. $N_b$ is varied from 4 to 10 with an interval of 2. BLMOL has higher accuracy in all cases compared to the simple stacked GCN baseline, which indicates that searched topology alleviates the over-smoothing problem by fusing features from different blocks. And in the NC task, a deeper GNN searched by BLMOL has a stronger representation ability. This phenomenon further illustrates the necessity of topology search. From Fig. \ref{fig_arch}, we observe that lower-level features are more likely to be exploited than higher-level features. This may be because high-level features obtained by multiple stacking become indistinguishable. Overall, by automatically searching for different topologies, BLMOL can effectively utilize features from different blocks to achieve better performance compared to existing handcrafted methods.

\begin{figure}
	\centering
	\subfloat[$P_c$]{
		\includegraphics[width=0.5\linewidth]{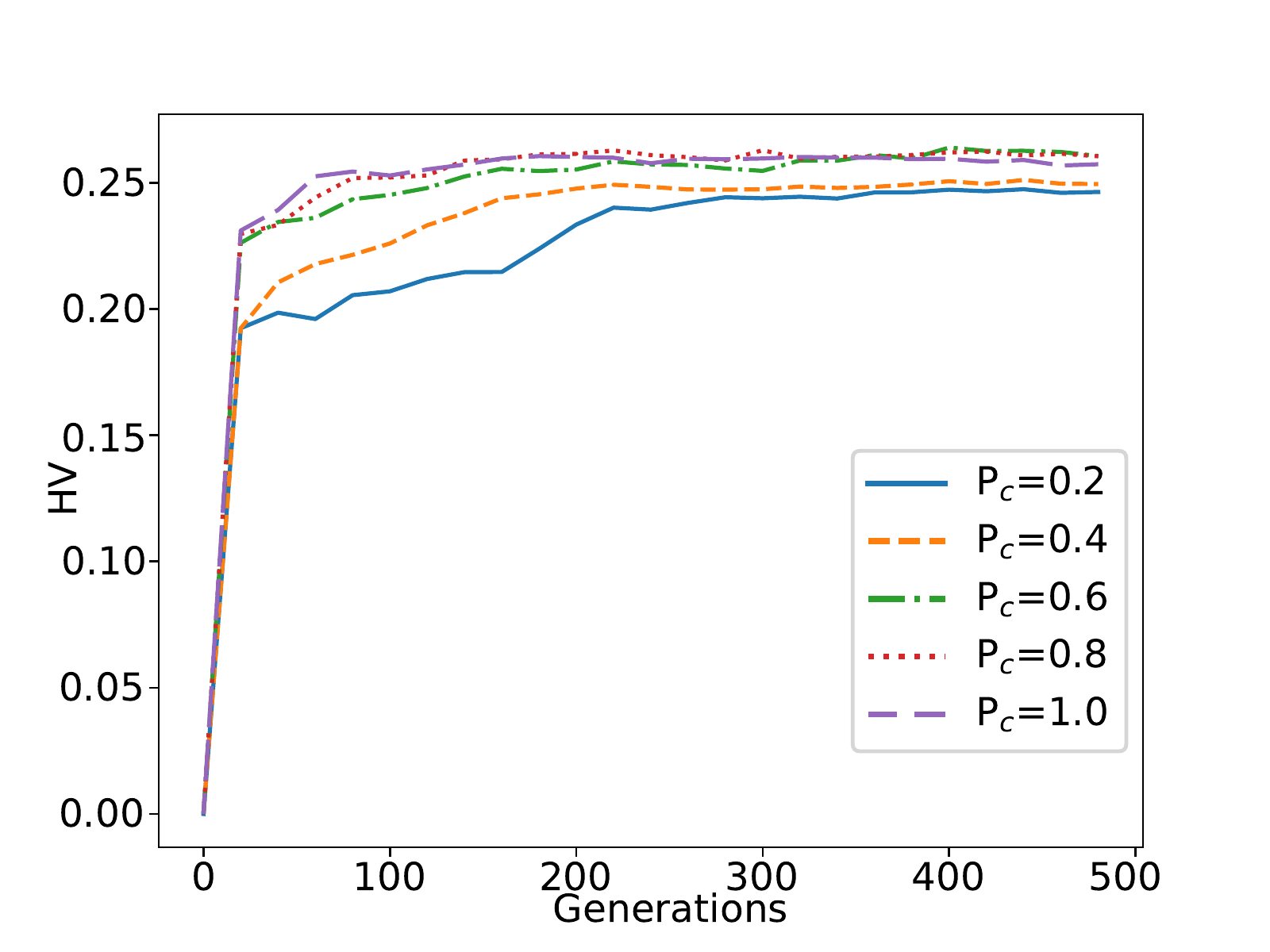}}
	\subfloat[$P_m$]{
		\includegraphics[width=0.5\linewidth]{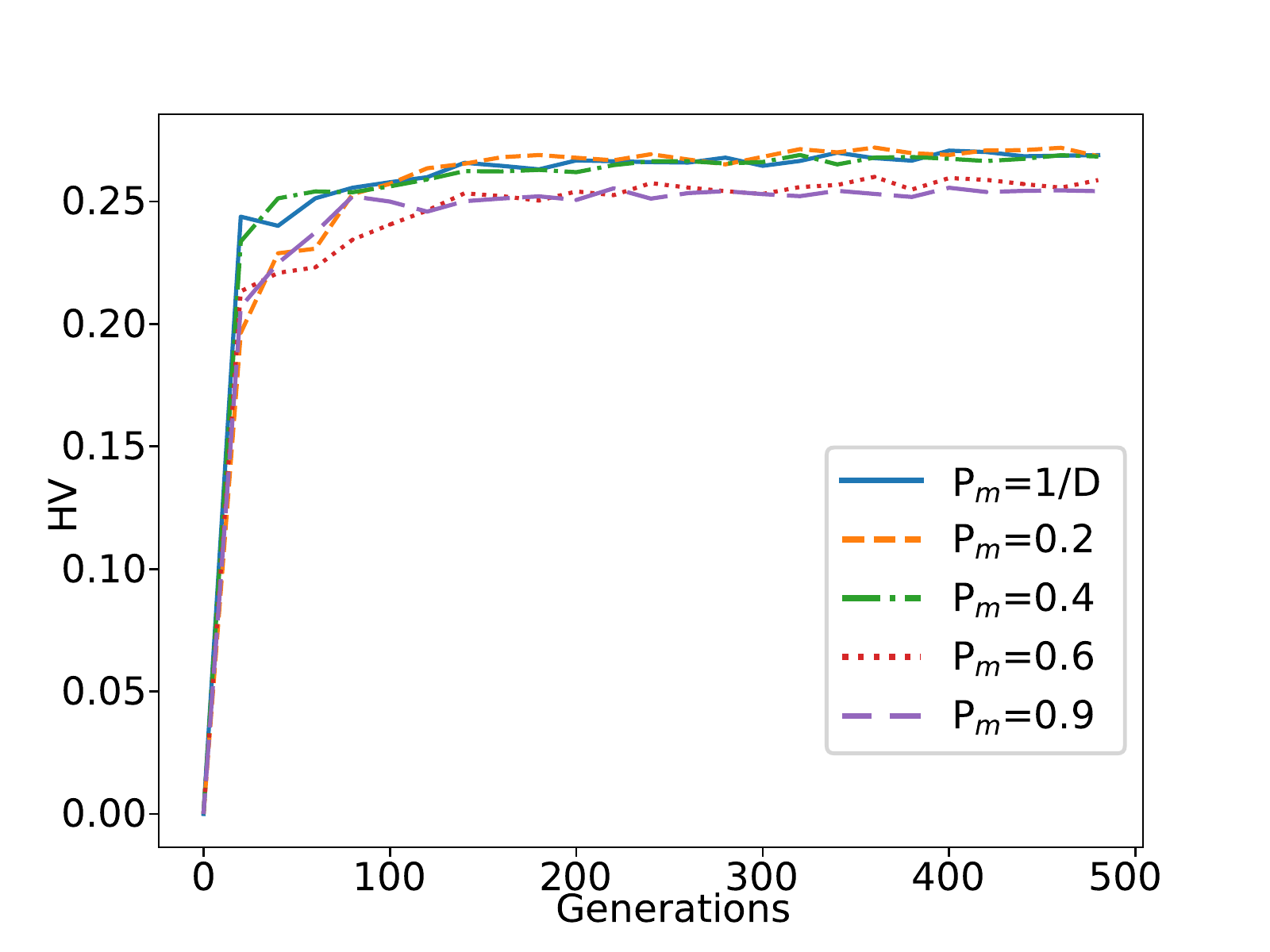}}
	\caption{Effects of crossover probability $P_c$ and mutation probability $P_m$ on the average (over 15 independent runs) HV.}
	\label{PCPM}
\end{figure}

In this paper, NSGAII is embedded into BL-MOL to optimize multiple UL objectives. The crossover and mutation operators used in NSGAII are accordingly controlled by two important hyperparameters: crossover probability $P_c$ and mutation probability $P_m$. To identify a good hyperparameter configuration, we performed the following parameter analysis experiments. We sweep the value of $P_c$ from 0.2 to 1 and $P_m$ from $1/D$ to 0.9, while setting the rest of the hyperparameters to their default values (see Table \ref{Parameter-settings}). For each setting, we run NSGAII 15 times on the ENZYMES datasets to maximize multiple UL surrogate objectives. The aggregation operation is set to GCN. HV\cite{797969}, a widely used metric in the field of MOO, is employed as the evaluation metric. Fig. \ref{PCPM} demonstrates the effects of crossover probability $P_c$ and mutation probability $P_m$ on the average (over 15 independent runs) HV. It can be observed that an increase in $P_c$ and a decrease in $P_m$ have a positive impact on performance. Therefore, in our experiments, the crossover probability and mutation probability are set to 1 and $1/D$, respectively. These hyperparameter configurations are also widely used in existing MOEAs\cite{deb2002fast,4358754}.

\section{Conclusion}
In this paper, we investigate how to solve the BL-MOP in machine learning. A surrogate-assisted bi-level multi-objective learning framework is proposed, which couples the LL decision-making process to the UL optimization process to overcome the computational challenges caused by the Pareto optimality of the sub-problems. Our proposed BLMOL can obtain multiple machine learning systems with different trade-offs among objectives. To demonstrate the effectiveness of the BLMOL framework, a series of experiments are carried out. Multi-task graph neural topology search is considered an illustrative case study. Experimental results on several graph machine learning tasks in three real-world datasets show that BLMOL can find a set of well-representative topologies and their weights. And GNNs obtained by BLMOL outperform the existing handcrafted ones.

In BLMOL, the quality of the obtained Pareto set is affected by the accuracy of the preference surrogate model. This makes BLMOL perform poorly for BL-MOPs which own an insufficient evaluation budget or a large UL search space. In the future, we plan to extend the BLMOL framework to more BL-MOPs in machine learning, such as feature selection\cite{7339682,9641743}, federated learning \cite{8744465,https://doi.org/10.48550/arxiv.2210.08295}, and fair learning\cite{9902997,https://doi.org/10.48550/arxiv.2207.12138}. In these paradigms, there often exist multiple objectives that need to be optimized simultaneously. Notably, since BLMOL makes no assumptions about the problem to be solved, it can be well generalized. According to the specific form of the objective, a series of potential improvements for specific applications need to be further studied in detail. Furthermore, we will investigate the relationship between multi-task graph topology and graph properties. Multi-task neural topology search on large-scale graphs is also a promising direction.


\bibliographystyle{IEEEtran}
\bibliography{my}





\vfill

\end{document}